\documentclass[10pt,twocolumn,letterpaper]{article}

\usepackage[pagenumbers]{iccv} %

\newcommand{\TODO}[1]{\textbf{\color{red}[TODO: #1]}}
\renewcommand{\TODO}[1]{}

\usepackage{subcaption}
\usepackage{multirow}
\usepackage[most]{tcolorbox}
\usepackage{fontawesome}

\def\eg{\emph{e.g}\onedot} 

\def\ie{\emph{i.e}\onedot}

\definecolor{mygreen}{RGB}{0,200,0}
\newcommand{\perfdown}[1]{\bf\textcolor{red}{#1}}
\newcommand{\perfup}[1]{\bf\textcolor{mygreen}{#1}}

\makeatletter
\let\@oldmaketitle\@maketitle%
\renewcommand{\@maketitle}{\@oldmaketitle%
     \centering
     \vspace{-.5em}
    \includegraphics[width=1.0\linewidth]{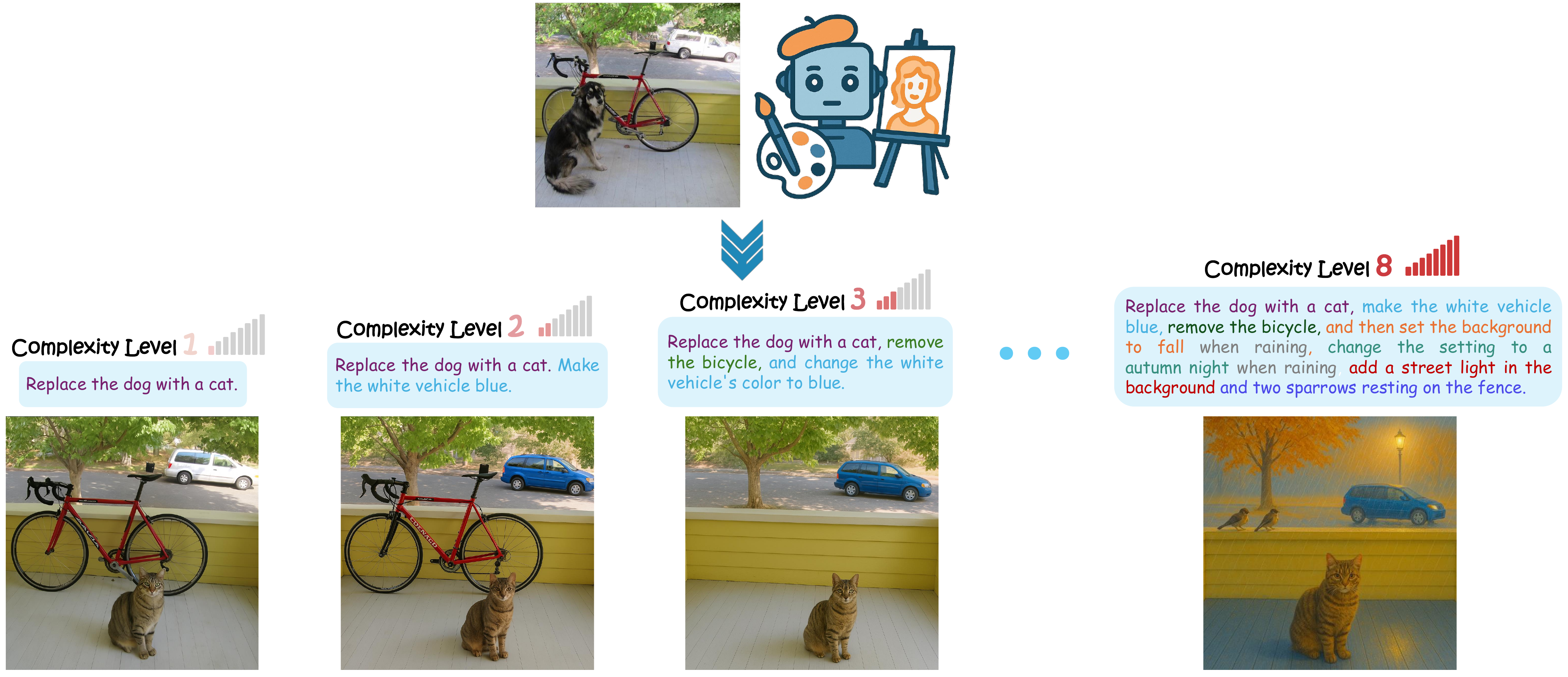}
    \vspace{-1.5em}
     \captionof{figure}{An illustration of our \texttt{Complex-Edit} Benchmark. This figure presents a structured progression of instruction complexity in image editing tasks, highlighting the transition from atomic edits to highly intricate transformations.}
    \label{fig:teaser}
    \bigskip}%
\makeatother

\definecolor{iccvblue}{rgb}{0.21,0.49,0.74}
\usepackage[pagebackref,breaklinks,colorlinks,allcolors=iccvblue]{hyperref}
\usepackage[capitalize]{cleveref}
\crefname{section}{Sec.}{Secs.}
\Crefname{section}{Section}{Sections}
\Crefname{table}{Table}{Tables}
\crefname{table}{Tab.}{Tabs.}
\Crefname{figure}{Figure}{Figures}
\crefname{figure}{Fig.}{Figs.}

\def\paperID{12818} %
\def\confName{ICCV}
\def\confYear{2025}

\title{\texttt{Complex-Edit}: CoT-Like Instruction Generation for Complexity-Controllable Image Editing Benchmark}

\author{%
  Siwei Yang$^{1}$ \, 
  Mude Hui$^{1}$ \,
  Bingchen Zhao$^{2}$ \,
  Yuyin Zhou$^{1}$ \,
  Nataniel Ruiz$^{3}$ \,
  Cihang Xie$^{1}$ \vspace{.3em}\\ 
  $^1$ University of California, Santa Cruz \quad $^2$ University of Edinburgh \quad
  $^3$ Google \\
  \\
  \small
  \hspace{3em} \includegraphics[height=1.1em]{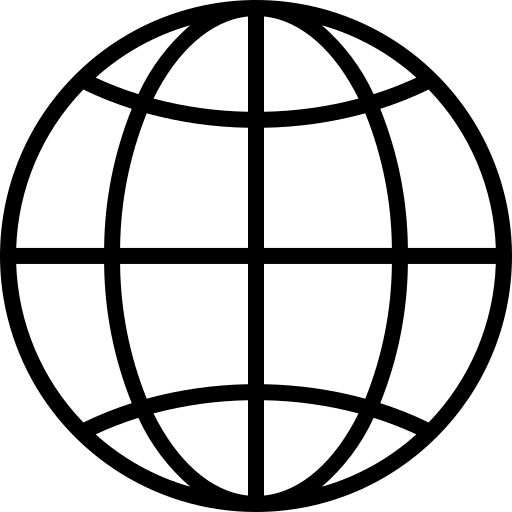} \textbf{Project Page}: \url{https://ucsc-vlaa.github.io/Complex-Edit/} \\
  \small
  \hspace{3em} \includegraphics[height=1.5em]{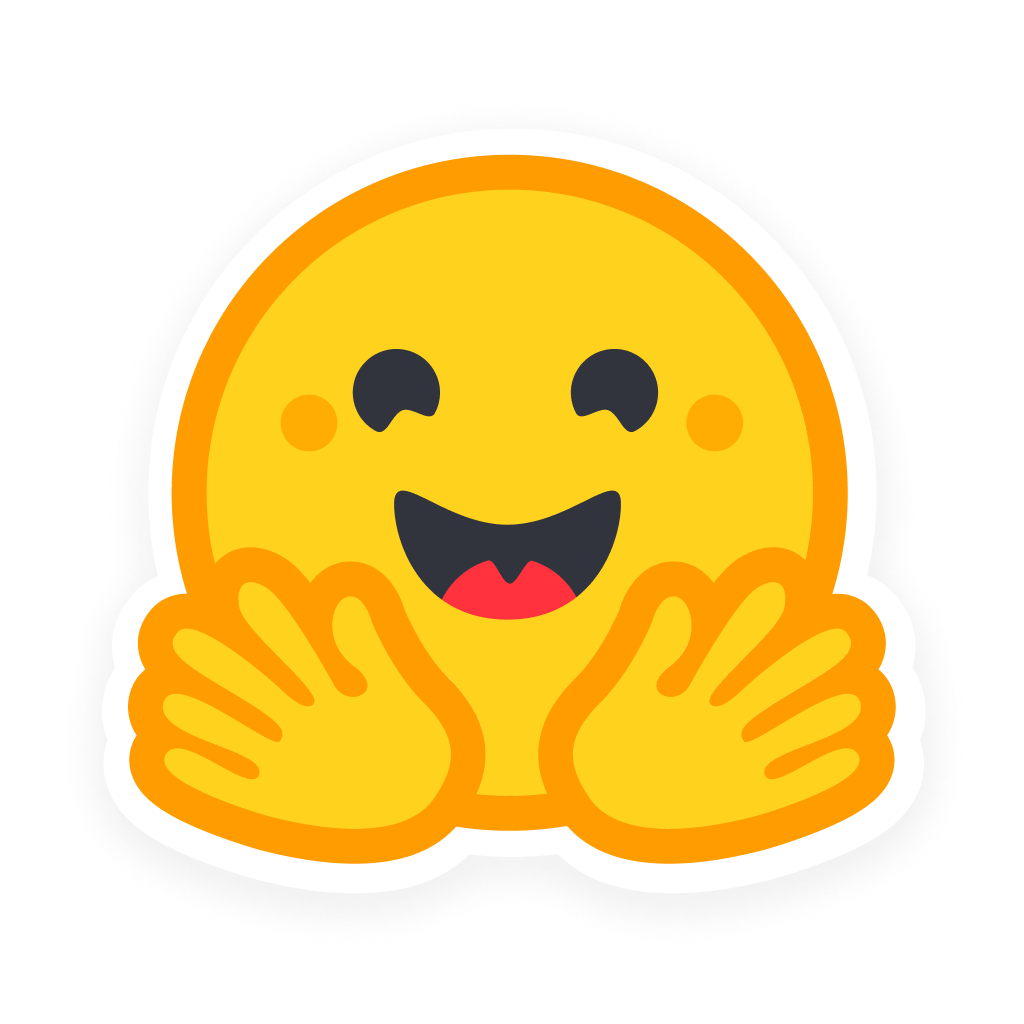} \textbf{Dataset}: \url{https://huggingface.co/datasets/UCSC-VLAA/Complex-Edit} \\
}

\begin{document}

\maketitle

\begin{abstract}
We introduce \texttt{Complex-Edit}, a comprehensive benchmark designed to systematically evaluate instruction-based image editing models across instructions of varying complexity. To develop this benchmark, we harness GPT-4o to automatically collect a diverse set of editing instructions at scale. Our approach follows a well-structured ``Chain-of-Edit'' pipeline: we first generate individual atomic editing tasks independently and then integrate them to form cohesive, complex instructions. 
Additionally, we introduce a suite of metrics to assess various aspects of editing performance, along with a VLM-based auto-evaluation pipeline that supports large-scale assessments.

Our benchmark yields several notable insights: 
1) Open-source models significantly underperform relative to proprietary, closed-source models, with the performance gap widening as instruction complexity increases;
2) Increased instructional complexity primarily impairs the models’ ability to retain key elements from the input images and to preserve the overall aesthetic quality; 
3) Decomposing a complex instruction into a sequence of atomic steps, executed in a step-by-step manner, substantially degrades performance across multiple metrics;
4) A straightforward Best-of-N selection strategy improves results for both direct editing and the step-by-step sequential approach; and
5) We observe a ``curse of synthetic data'': when synthetic data is involved in model training, the edited images from such models tend to appear increasingly synthetic as the complexity of the editing instructions rises --- a phenomenon that intriguingly also manifests in the latest GPT-4o outputs.
\end{abstract}
    
\section{Introduction}
\label{sec:intro}

\begin{figure*}[t!]
    \centering
    \centering
    \includegraphics[width=\linewidth]{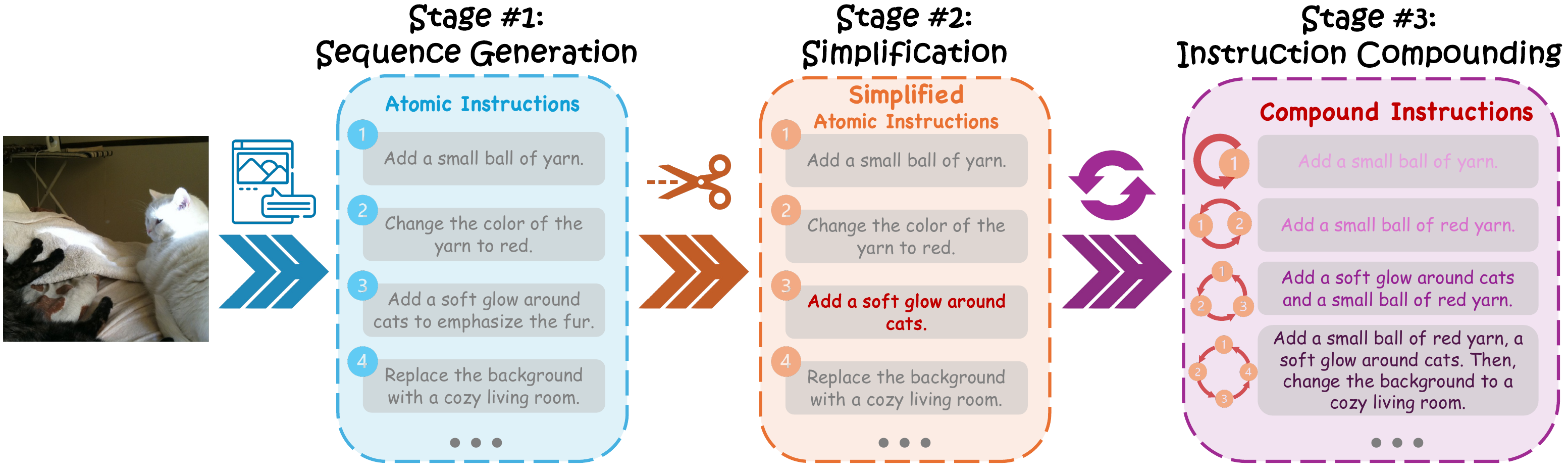}
    \vspace{-2em}
    \caption{An overview of our data collection pipeline. The pipeline consists of three distinct stages: 1) \textbf{Stage \#1 Sequence Generation}: for each image, a series of atomic instructions is produced; 2) \textbf{Stage \#2 Simplification}: each fundamental instruction is refined to eliminate extraneous details, preserving only the essential description of the editing process; 3) \textbf{Stage \#3 Instruction Compounding}: several atomic instructions are integrated into one comprehensive instruction.}
    \vspace{-1em}
    \label{fig:data_pipeline_example}
\end{figure*}

Since InstructPix2Pix~\cite{brooks2023instructpix2pix} introduced instruction-based image editing---where images are directly modified through textual commands---the field has witnessed remarkable progress.
Noteworthy developments include enhancements in the quality of training images \cite{kawar2023imagic,zhang2024hive,hui2025hqedit,zhao2024ultraedit,wei2025omniedit,baldridge2024imagen}, the evolution of model architectures from CNN-based to transformer-based diffusion models \cite{sit}, the refinement of training methodologies \cite{simsar2024uip2p,shi2024seededit}, and the on-going exploration of test-time scaling \cite{guo2025can,ma2025inference}.

Despite these strides,  much of the available training data and evaluation benchmarks rely on relatively simple editing instructions~\cite{brooks2023instructpix2pix,zhang2023magicbrush,zhang2024hive,zhao2024ultraedit,yu2024anyedit,wei2025omniedit,sheynin2023emu,ku2024imagenhub}. Yet, real-world scenarios often demand the ability to handle instructions that vary considerably in complexity. For instance, while one user might simply request to ``remove the car'', another might need a more elaborate transformation, such as ``replace the car with a blue bus featuring an iPhone advertisement on its side''. The current lack of evaluation benchmarks capable of handling such diverse instruction complexities not only impedes our ability to rigorously assess existing editing models but also hampers the evaluation of emerging, more powerful systems~\cite{guo2025can,ma2025inference}.

To bridge this gap, we present a novel data generation pipeline that leverages advanced GPT models to create a diverse, scalable, and complexity-controllable evaluation dataset for image editing (see~\cref{fig:data_pipeline_example} for an illustration). 
Our approach follows a chain-of-thought-like~\cite{wei2022chain} paradigm, unfolding in three key stages.  First, in the \textit{Sequence Generation} stage, an input image prompts GPT-4o to generate a series of simple instructions corresponding to predefined atomic operations (\eg, ``Add an Object'', ``Change the Object Color''), with each serving as an intermediate step toward a more complex editing task.
Next, recognizing that these GPT-generated atomic instructions may contain superfluous/unnecessary details---such as added commentary on the operation's intention---we then move to a \textit{Simplification} stage, where each instruction is trimmed to retain only its core editing intent.
In this final \textit{Instruction Compounding}  stage, the simplified atomic instructions are merged into a single, coherent complex instruction (again, via GPT-4o), using the input image as contextual guidance. We name this new dataset \texttt{Complex-Edit}. Importantly, this structured process of creating \texttt{Complex-Edit} allows us to quantitatively control the editing instruction’s complexity by simply adjusting the number of merged atomic instructions.

To evaluate performance on these complex instructions, we developed a comprehensive evaluation framework that measures three critical dimensions: 
(1) \emph{Instruction Following}, which assesses whether the intended modifications are correctly applied; 
(2) \emph{Identity Preservation}, which ensures that unspecified elements remain unchanged; 
and (3) \emph{Perceptual Quality}, which evaluates the overall aesthetic quality and absence of artifacts.
To support large-scale evaluations, we implement a vision language model (VLM)-based autograding system to quantify these dimensions. 
Our investigation further identifies important considerations in enhancing this evaluation pipeline.
For example, enabling chain-of-thought reasoning in VLM evaluations does not necessarily enhance evaluation quality---contrary to previous findings~\cite{ku2024viescore,hui2025hqedit}---while providing detailed rubrics and using direct numeric scoring~\cite{xie2023self,xie2024monte,zhou2024calibrated,cui2024fine} consistently improves autograder performance.

Our extensive evaluation results reveal five key insights that were previously difficult to capture with existing benchmarks.
First, open-source models significantly underperform compared to proprietary, closed-source models, with this disparity growing as editing complexity increases. 
Second, the increasing instruction complexity mainly affects editing models' perceptual quality (especially in identity preservation), while its impact on instruction following varies across models. 
Thirdly, unlike the benefits observed with chain-of-thought (CoT) reasoning in text generation, CoT-inspired sequential editing yields much worse results than directly executing complex instructions --- we note this is because sequential editing significantly degrades outputs in instruction following, identity preservation, and perceptual quality.
Fourth, simple techniques such as Best-of-N sampling improve direct editing outcomes across all three dimensions and, when applied to sequential editing, notably enhance identity preservation and perceptual quality. Lastly, our \texttt{Complex-Edit} reveals an emerging `curse of synthetic data' --- when training utilizes synthetic images, models tend to produce edited outputs with overly synthetic qualities under complex instructions, often resembling oil paintings or animations. More interestingly, this phenomenon is also observed with the latest GPT-4o models, potentially suggesting that the incorporation of synthetic data may contribute to GPT-4o’s advanced image generation capabilities.

We hope this newly developed benchmark, \texttt{Complex-Edit}, will not only deepen the community’s understanding of instruction-based image editing models but also serve as a valuable framework for the rigorous evaluation of next-generation image editing systems, particularly those with test-time scalability.

\section{Collection of \texttt{Complex-Edit}}
\label{sec:data}
As illustrated in \cref{fig:data_pipeline_example}, the collection of \texttt{Complex-Edit} is organized into three stages: 1) \textbf{Stage \#1 Sequence Generation}: a sequence of atomic instructions are generated for each image; 2) \textbf{Stage \#2 Simplification}: each atomic instruction is simplified to remove unnecessary information other than the description of the editing operation; 3) \textbf{Stage \#3 Instruction Compounding}: multiple atomic instructions are combined into a single, complex instruction.

\subsection{Sequence Generation}
\label{subsec:data_sequence}
\begin{figure*}[t!]
    \centering
    \centering
    \includegraphics[width=\linewidth]{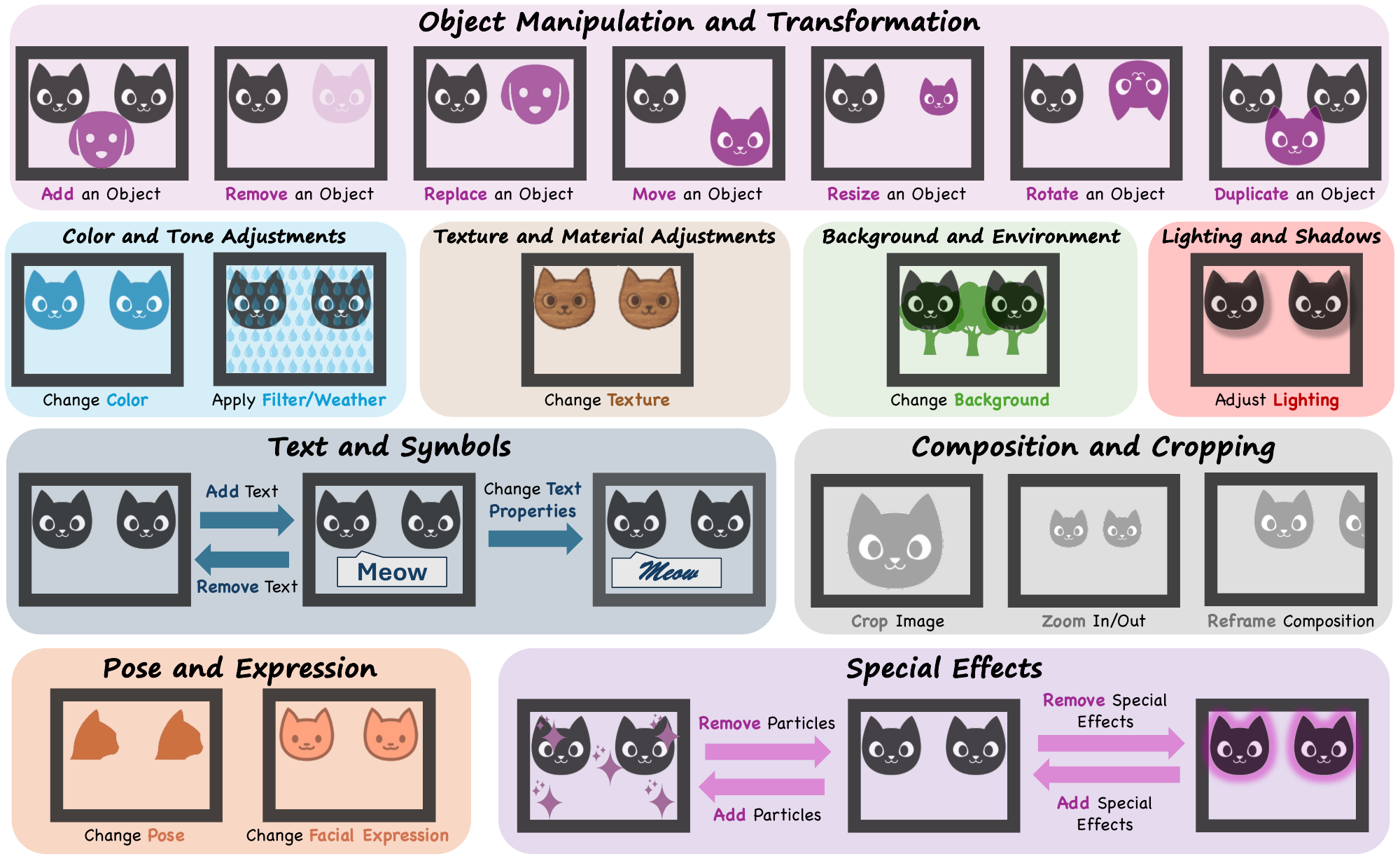}
    \vspace{-2em}
    \caption{An illustration of 24 types of atomic editing operations in 9 categories.}
    \vspace{-1.1em}
    \label{fig:edit_teaser}
\end{figure*}

In this first stage, we begin by defining 24 distinct atomic operations that represent the most basic actions of image editing. These operations can be grouped into 9 categories --- \textit{Object Manipulation and Transformation}, \textit{Color and Tone Adjustments}, \textit{Texture and Material Adjustments}, \textit{Background and Environment, Lighting and Shadows}, \textit{Text and Symbols}, \textit{Composition and Cropping}, \textit{Pose and Expression}, \textit{Special Effects} --- each of which captures a unique aspect of image editing, as shown in Fig.~\ref{fig:edit_teaser}.

\begin{figure}[t!]
    \centering
    \vspace{-.5em}
    \includegraphics[width=\linewidth]{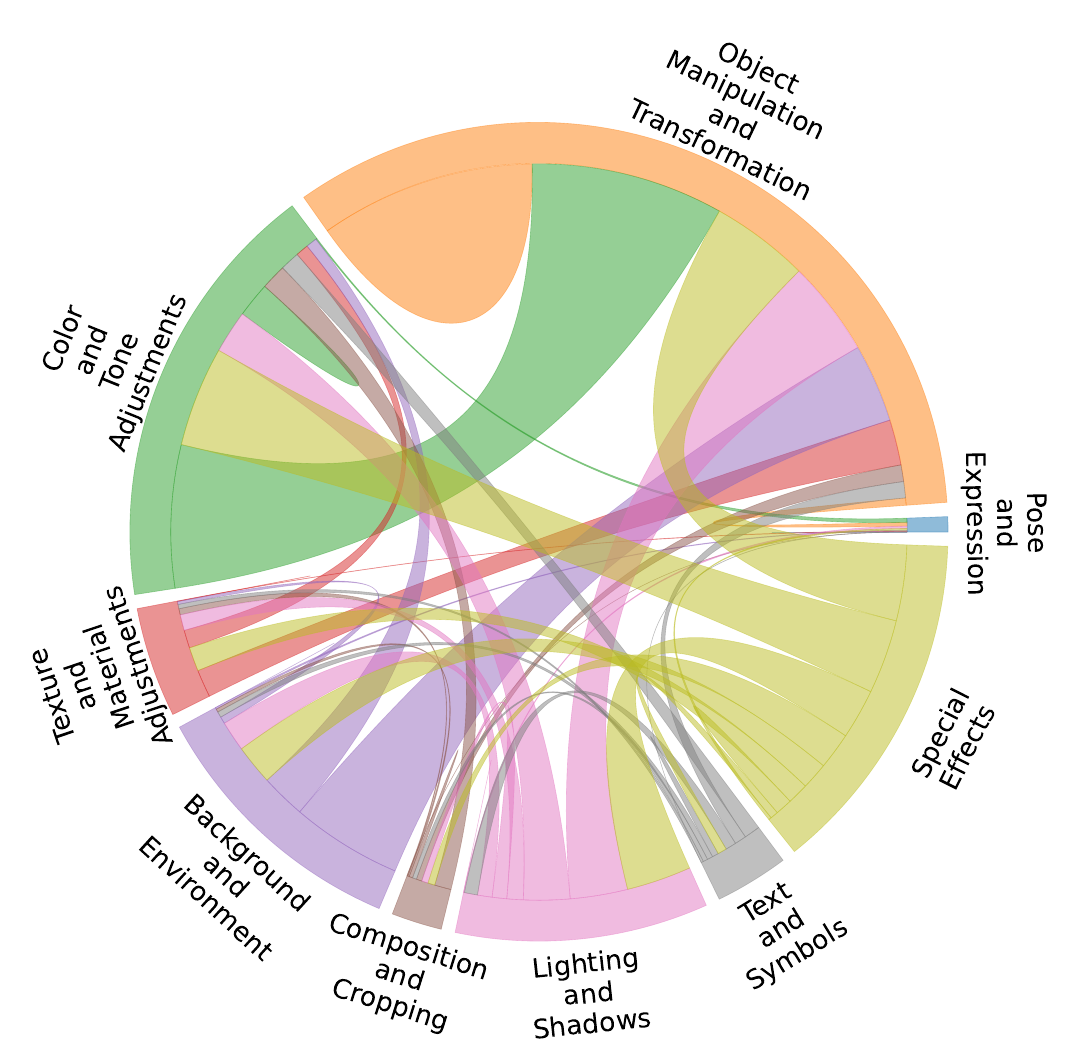}
    \vspace{-2em}
    \caption{The distribution of atomic instructions among 9 operation categories. The arc thickness between two categories shows the number of adjacent instructions from these two categories.}
    \vspace{-1em}
    \label{fig:dataset_stastics}
\end{figure}

We provide detailed descriptions of each operation type in the supplementary materials~\cref{sec:supp_atomic}.
These categorized atomic operation types and their descriptions, along with an input image, are fed into GPT-4o to generate a sequence of atomic instructions. Note that each generated instruction adheres to a predetermined length and is associated with one of the predefined atomic operations.
We also record the atomic operation type for each instruction, with its distribution illustrated in \cref{{fig:dataset_stastics}}. This metadata can facilitate the future development of specialized image editing models~\cite{wei2025omniedit,yu2024anyedit}.

\subsection{Simplification}
\label{subsec:data_simplify}
In practice, these GPT-generated instructions sometimes are unnecessarily rich in detail and may include extraneous commentary on the intentions of the editing operations, as shown in \cref{fig:data_pipeline_example}. Because our benchmark is designed to evaluate the performance of image-editing models on clear and concise instructions, 
we employ a dedicated simplification stage to eliminate any superfluous content. Specifically, each generated atomic instruction is examined by GPT-4o to determine whether it requires further simplification; if so, GPT-4o outputs a simplified version adhering to the predefined response format, and the original instruction is replaced accordingly.

\subsection{Instruction Compounding}
Lastly, we combine these simplified atomic instructions into compound instructions. Specifically, given a sequence of $N$ atomic instructions, we progressively generate $N$ compound instructions corresponding to different levels of complexity: a compound instruction at complexity level $C_i$ is formed by combining the first  $i$ atomic instructions in the sequence, with the simplest level ($C_1$) being identical to the first atomic instruction and the hardest level ($C_N$) integrating all $N$ atomic instructions. 

In implementation, rather than simply concatenating the instructions, GPT-4o is employed to seamlessly integrate them, potentially reordering or merging operations to make the final outcome more natural and coherent. For example, as shown in \cref{fig:data_pipeline_example}, rather than sequentially executing ``add a ball of yarn'' and ``change the color of the yarn to red'', the instructions are compounded into a single directive:  ``add a red ball of yarn''.

\subsection{Implementation Details}
To ensure that GPT-4o fully understands our objectives, we carefully construct multiple few-shot examples for each of the three stages. Additionally, CoT reasoning~\cite{wei2022chain} is enabled in both Stage \#1 (Sequence Generation)  and Stage \#3 (Instruction Compounding) to enhance generation quality. Furthermore, to improve data diversity, we slightly increase the sampling temperature from 1.0 to 1.15 during Stage \#1 (Sequence Generation). Although this adjustment promotes diversity, it also increases the likelihood of generating garbled text. To address this, we implement a rule-based filtration mechanism at each stage that detects and regenerates any flawed outputs.

\section{Evaluation}
\label{sec:metric}
Following prior works \cite{ku2024viescore,hui2025hqedit}, we employ VLM-based autograders to facilitate large-scale evaluation. However, we have observed that these frameworks do not fully capture the nuances of instruction-guided image editing. In response, we introduce several enhancements to address these limitations and improve the overall autograding framework.

\subsection{Metric Design}
Our evaluation framework focuses on two primary dimensions: 1) the alignment dimension---whether the output image reflects the changes specified by the editing instruction, and 2) the perceptual quality dimension---whether the output image looks aesthetically pleasing and is devoid of visual artifacts. 

\begin{figure*}[th]
    \centering
    \centering
    \includegraphics[width=\linewidth]{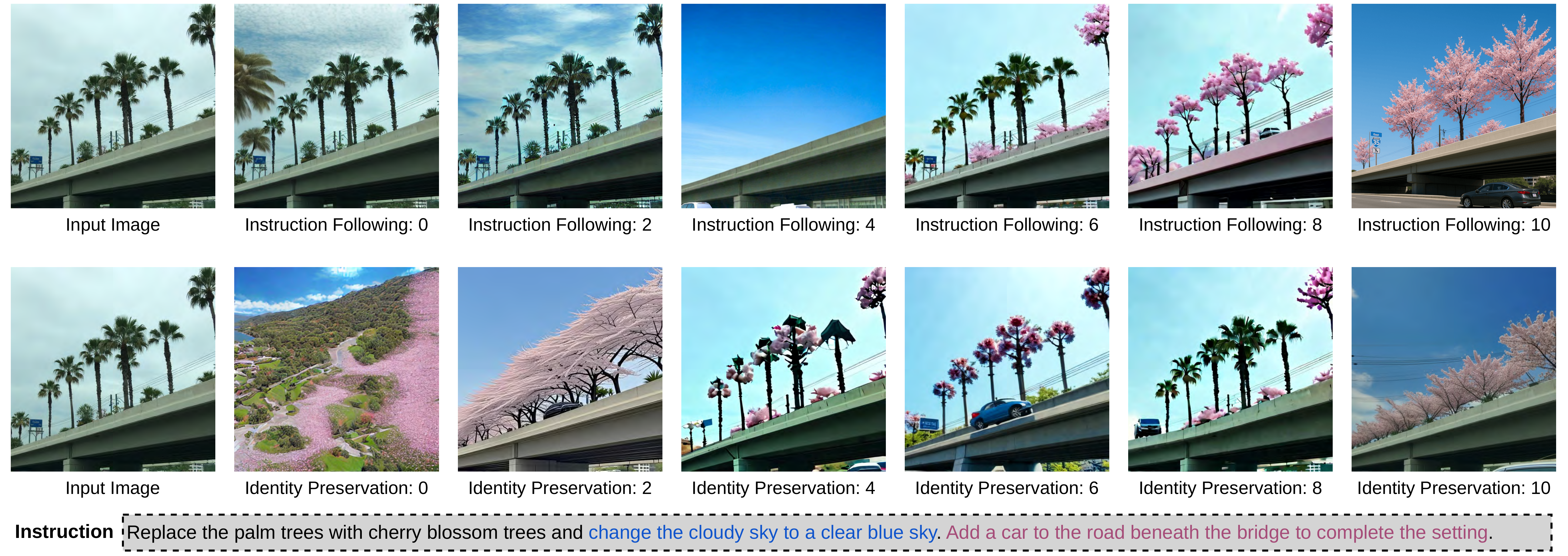}
    \vspace{-2em}
    \caption{Examples of evaluation results for Instruction Following and Identity Preservation.}
    \vspace{-.5em}
    \label{fig:alignment}
\end{figure*}

\begin{figure*}[th]
    \centering
    \centering
    \includegraphics[width=\linewidth]{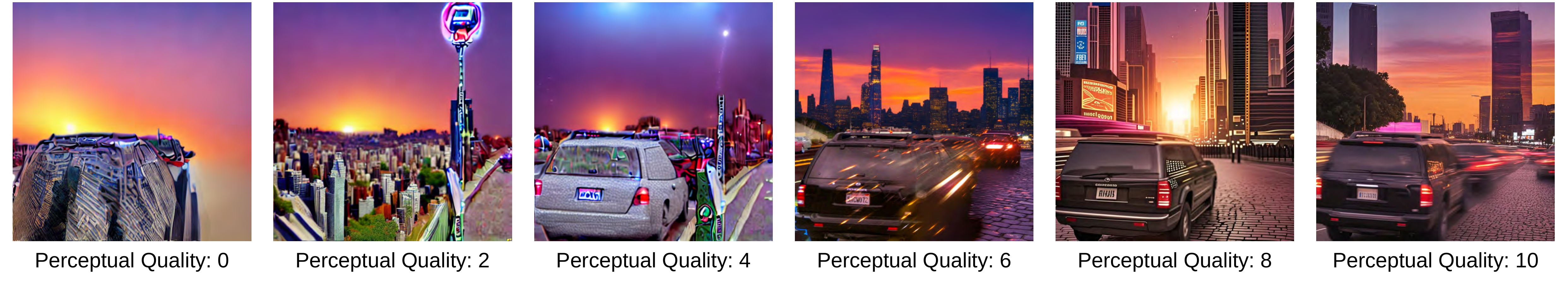}
     \vspace{-2.2em}
    \caption{Examples of evaluation results for Perceptual Quality.}
    \vspace{-.5em}
    \label{fig:quality}
\end{figure*}

\subsubsection{Alignment}
Unlike HQ-Edit~\cite{hui2025hqedit}, which simply measures alignment as a single metric, we hereby decompose it into two complementary criteria:

\begin{itemize}
    \item \textbf{Instruction Following} ($IF$): 
    Measures whether the specified modifications are present in the output image.
    \item \textbf{Identity Preservation} ($IP$):
    Assesses whether elements of the input image that should remain unchanged are indeed preserved.
\end{itemize}

We would like to point out that these two criteria roughly correspond to the directional CLIP similarity ($\text{CLIP}_\text{dir}$) and image-wise CLIP similarity ($\text{CLIP}_\text{img}$) metrics used in earlier studies~\cite{brooks2023instructpix2pix,zhang2023magicbrush,zhang2024hive}. 
When accessing $IF$ and $IP$, we feed the VLM with a triplet containing \{\textit{an input image}, \textit{an output image}, and \textit{an editing instruction}\}. For better visual understanding, we provide a series of example results of $IF$ and $IP$ in \cref{fig:alignment}.

\subsubsection{Perceptual Quality}
Beyond alignment, we also evaluate the overall visual quality of the generated images. Specifically, our \textbf{Perceptual Quality (PQ)} metric examines factors such as consistency in lighting and shadows, style coherence, and the seamless integration of elements.  
Because users may sometimes request bizarre or non-aesthetic edits, such as ``add motion blur to these vehicles'' or ``enlarge the pet to match the height of the owner'', one might argue that it is essential to include editing instructions when evaluating perceptual quality. However, as discussed in \cref{subsubsec:meta_eval_inst_input}, our empirical results indicate that providing editing instructions to VLMs actually reduces the correlation between VLM and human evaluations. Consequently, for $PQ$ evaluation, we only supply the VLM autograder with the edited image. Sample $PQ$ results are provided in \cref{fig:quality}.

\subsubsection{Overall Score}
Lastly, we define the overall score, $O$, simply as the arithmetic mean of the three metrics to summarize performance:

\begin{equation}
    O = \frac{IF + IP + PQ}{3}
\end{equation}

\subsection{Metric Calculation}
\subsubsection{Numeric Scoring v.s. Token Probability}
In line with VIEScore and HQ-Edit, we instruct the VLM to assign a score between 0 and 10 for each metric. We refer to this method as \textbf{numeric scoring}. In addition, we explore another approach for metric computation: \textbf{token probalility}, which is widely used in test-time scaling and VLM post-training~\cite{xie2023self,xie2024monte,zhou2024calibrated,cui2024fine}. Specifically, rather than directly asking the VLM for a numerical score, we reformulate the evaluation as a binary classification task by posing a \textit{yes-or-no question} to the VLM, \eg, ``Do the specified changes appear in the output image?''. The score is then calculated as the normalized token probability for the affirmative response ``Yes'' using $\text{Prob}_\text{Yes} / (\text{Prob}_\text{Yes} + \text{Prob}_\text{No})$.

\subsubsection{Detailed Rubric}
In contrast to existing frameworks like VIEScore and HQ-Edit, we design comprehensive rubrics for each metric to guide the evaluation process and improve result interpretability. Detailed descriptions of these rubrics are provided in the supplementary material \cref{subsec:supp_prompt_rubrics}. In addition, we discuss the impact of these rubrics and different scoring approaches in \cref{subsubsec:meta_eval_scoring}.

\subsection{Per-sample Variance}
\label{subsec:metric_variance}
Previous evaluations of image-editing models using advanced proprietary VLMs, such as GPT-4o \cite{hurst2024gpt} and Gemini \cite{team2023gemini,team2024gemini}, have encountered issues with score stability, leading to score variations for individual samples across different runs. Although averaging over many samples generally mitigates these discrepancies at the dataset level, per-sample stability is crucial for test-time scaling, where accurate and consistent evaluation is required for each individual sample~\cite{xie2024monte,guo2025can,ma2025inference}.

A straightforward approach to enhance stability utilizes multiple measurements to average the resultant scores, though this approach may considerably escalate computational costs.
Alternatively, one can attempt to make VLM outputs more deterministic by using greedy random sampling techniques (\eg, by setting the sampling probability mass to an extremely low value such as 1e-7). However, empirically, this deterministic approach slightly lower the correlation with human evaluations as shown in \cref{tab:deterministic}.  Consequently, we adopted the first approach by default, utilizing the average score from 20 independent evaluations for each sample. Further discussion on reducing per-sample variance is included in \cref{subsec:meta_eval_variance}.

\section{Meta-Evaluation for Metrics}
\label{sec:meta_eval}

In this section, we detail our meta-evaluation of the VLM-based auto-evaluation framework, focusing on design choices and their effectiveness in assessing image editing model outputs. We expect an effective auto-evaluation pipeline to satisfy two core criteria: 1) High correlation with human judgments, and 2) Low variance for ensuring reproducibility. Next, we first examine how various design decisions impact correlation with human evaluations (Section~\ref{subsec:meta_eval_corr}) and then describe our strategies for reducing per-sample variance (Section~\ref{subsec:meta_eval_variance}).

\begin{table}[t!]
    \centering
    \begin{tabular}{@{}lccc@{}}
        \toprule
        \multirow{2}{*}{Deterministic} & \multicolumn{3}{c}{Correlation} \\
        \cmidrule(lr){2-4}
        & $IF$ & $IP$ & $PQ$\\
        \midrule
        & \textbf{0.468} & \textbf{0.530} & \textbf{0.234} \\
        \checkmark & 0.461 & 0.507 & 0.215 \\
        \bottomrule
    \end{tabular}
    \vspace{-.8em}
    \caption{Random sampling vs Discriminative sampling.}
    \vspace{-.7em}
    \label{tab:deterministic}
\end{table}

\begin{table}
    \centering
    \resizebox{\linewidth}{!}{
        \begin{tabular}{@{}lccccc@{}}
            \toprule
            \multirow{2}{*}{Scoring Method} & \multirow{2}{*}{CoT} & \multirow{2}{*}{Rubric} & \multicolumn{3}{c}{Correlation} \\
            \cmidrule(lr){4-6}
            & & & $IF$ & $IP$ & $PQ$\\
            \midrule
            Token Prob &  &  & 0.411 & 0.366 & 0.136 \\
            Token Prob & \checkmark &  & 0.447 & 0.460 & 0.158 \\
            \hline
            Numeric &  &  & 0.434 & 0.450 & 0.207 \\
            Numeric &  & \checkmark & 0.446 & 0.451 & \textbf{0.234} \\
            Numeric & \checkmark & \checkmark & \textbf{0.468} & \textbf{0.530} & 0.208 \\
            \toprule
            &  &  & $\text{CLIP}_\text{dir}$ & $\text{CLIP}_\text{img}$ &  \\
            \midrule
            &  &  & 0.182 & 0.523 &  \\
            \bottomrule
        \end{tabular}
    }
    \vspace{-.8em}
    \caption{The results from different scoring methods. We can observe that w/ COT and w/ Rubric are the best for IF and IP, and w/o COT and w/ rubric are the best for PQ.}
    \vspace{-1em}
    \label{tab:scoring_method}
\end{table}

\begin{figure*}[th]
    \centering
    \includegraphics[width=.85\linewidth]{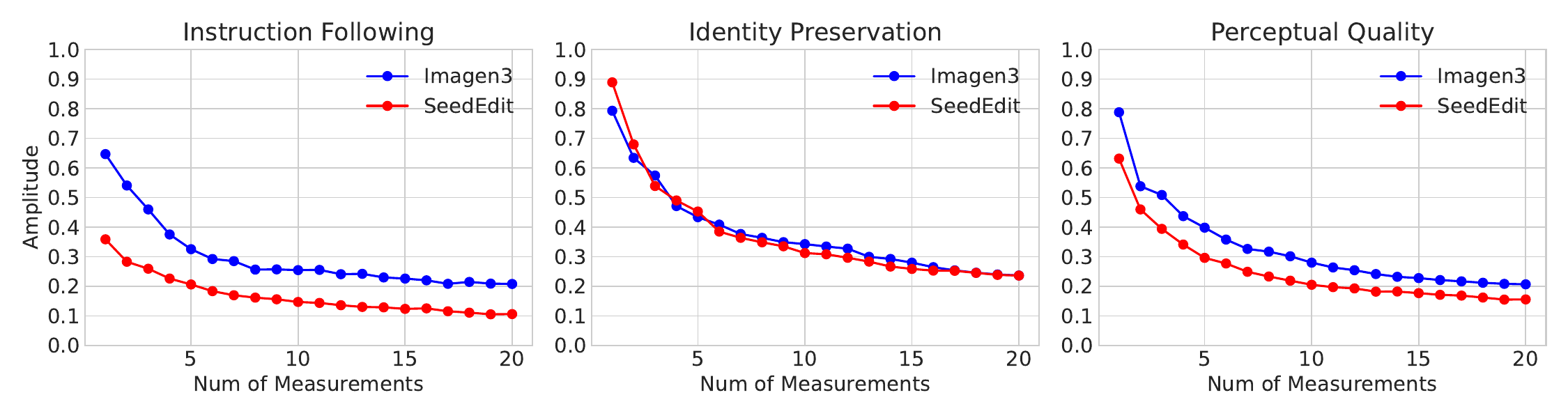}
    \vspace{-1em}
    \caption{Relationship between per-sample variance and the number of measurements}
    \label{fig:variance}
\end{figure*}

\begin{figure*}[t!]
    \centering
    \begin{subfigure}{\linewidth}
        \centering
        \includegraphics[width=\linewidth]{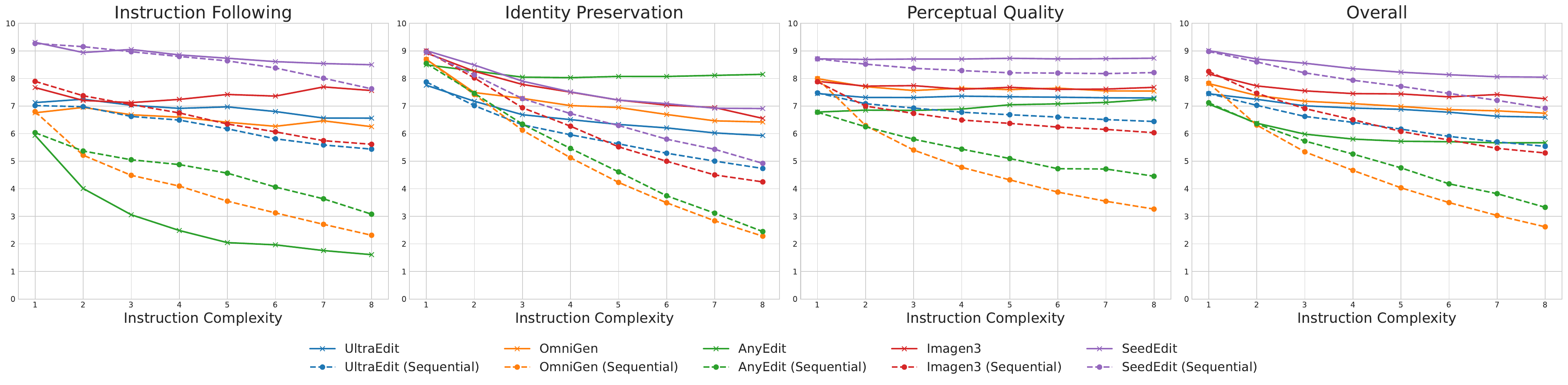}
        \caption{Real-life input images.}
        \label{fig:main_eval_real}
    \end{subfigure}
    \hfill
    \begin{subfigure}{\linewidth}
        \centering
        \includegraphics[width=\linewidth]{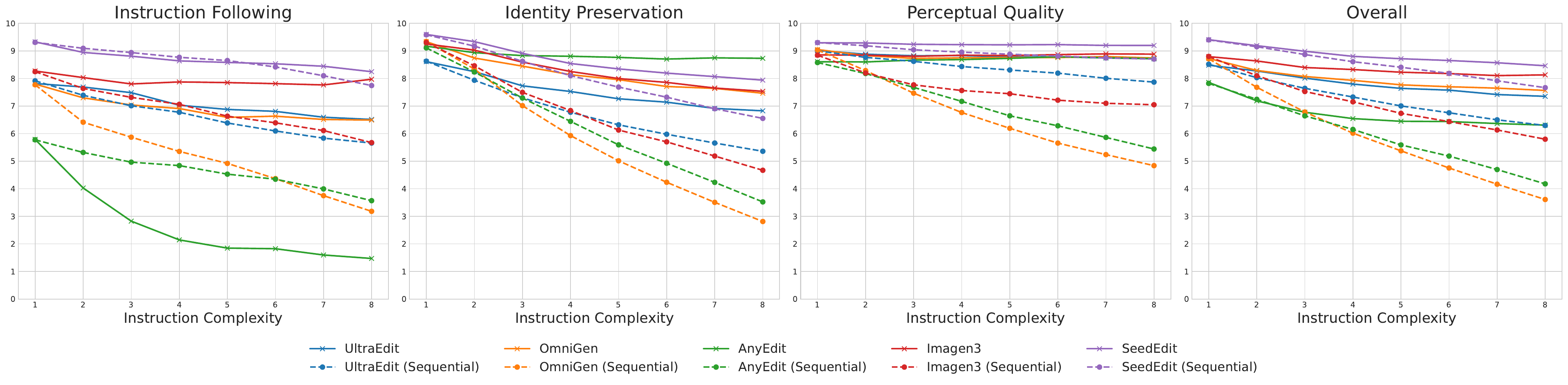}
        \caption{Synthetic input images.}
        \label{fig:main_eval_syn}
    \end{subfigure}
    \vspace{-2em}
    \caption{Evaluation results of direct and sequential editing. The instruction complexity level ranges from $C_1$ to $C_8$. This indicates that $IP$ and $PQ$ consistently dropped as the instruction complexity level grows, while it influence on $IF$ varies among different models. Also, the performance gap between models is enhanced with increasing complexity.}
    \label{fig:main_eval}
\end{figure*}

\begin{figure*}[ht!]
    \centering
    \begin{subfigure}{\linewidth}
        \centering
        \includegraphics[width=.99\linewidth]{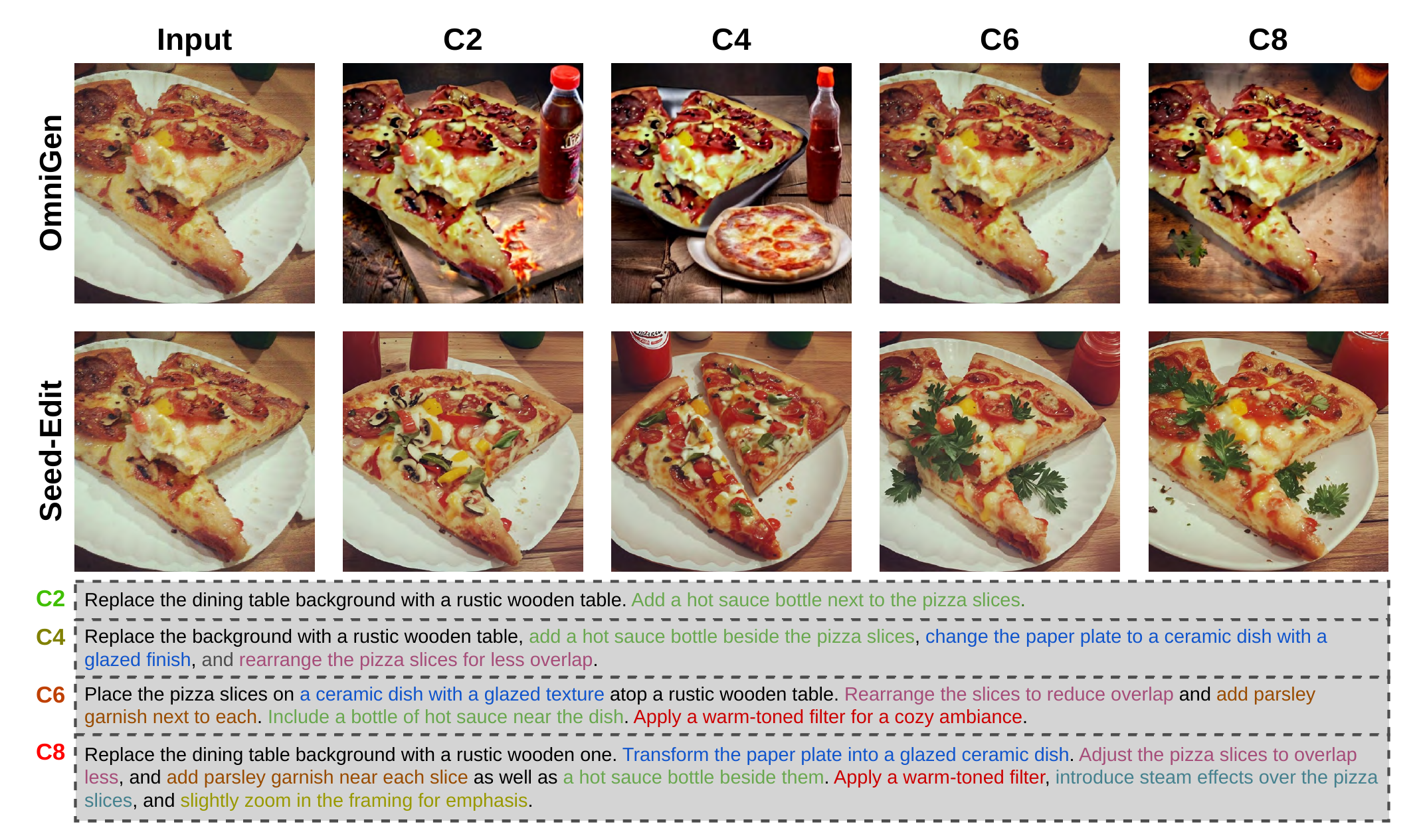}
        \vspace{-.5em}
        \caption{real input images.}
        \label{fig:vis_main_real}
    \end{subfigure}
    \hfill
    \vspace{-2em}
    \begin{subfigure}{\linewidth}
        \centering
        \includegraphics[width=.99\linewidth]{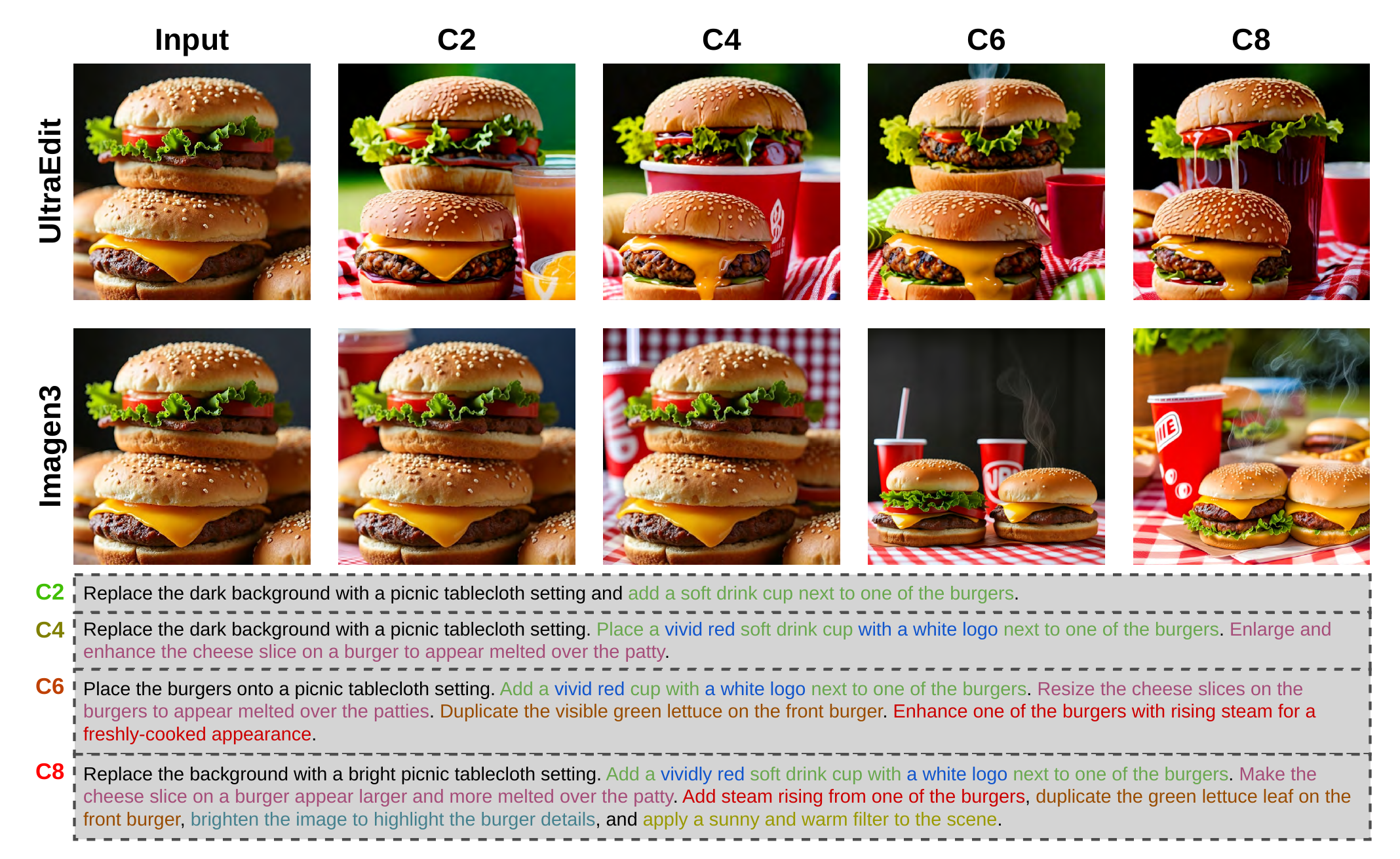}
        \vspace{-.5em}
        \caption{Synthetic input images.}
        \label{fig:vis_main_syn}
    \end{subfigure}
    \vspace{-1.5em}
    \caption{Outputs from open-source and proprietary models via direct editing. See results with more models in \cref{fig:vis_direct_real,fig:vis_direct_syn}.}
    \label{fig:vis_main}
\end{figure*}

\subsection{Correlation with Human Evaluation}
\label{subsec:meta_eval_corr}
\subsubsection{Meta-Evaluation Settings}
Meta-evaluation for instruction-based image editing metrics~\cite{ku2024viescore,hui2025hqedit} is typically performed by correlating human-assigned scores with metric values. However, we found that expecting human raters to assign consistent numeric scores across diverse samples is challenging. Instead, for each input image and editing instruction, we present raters with a pair of output images and ask them to compare the outputs with respect to each metric. We then compute the correlation between these human comparisons and the differences in the corresponding metric scores. Additionally, we ask raters to choose their preferred output based on overall impression. This comparative approach allows us to align our metric aggregation with human intuition and to determine the best way to combine the three proposed metrics into a single score.

To ensure that our meta-evaluations generalize across varying levels of instruction complexity, we sampled 100 output images for each complexity level from $C_1$ (\ie, the simplest level, just a single atomic instruction) to $C_8$ (\ie, the hardest level, compounding from 8 atomic instructions) using both Imagen3 \cite{baldridge2024imagen}  and SeedEdit \cite{shi2024seededit}, resulting in 800 image pairs. Each pair was annotated independently by at least two human raters. 
We use GPT-4o as our VLM evaluator. Since the per-sample variance is yet to be discussed in \cref{subsec:meta_eval_variance}, for now, we \textit{naively} evaluate each sample 40 times and average the scores to reduce the per-sample variance.

\subsubsection{Scoring Method and Rubric}
\label{subsubsec:meta_eval_scoring}
As shown in \cref{tab:scoring_method}, our experiments indicate that numeric scoring yields a higher correlation with human evaluation than token probability methods. Furthermore, supplementing the numeric scoring with a detailed rubric consistently improves this correlation, emphasizing the importance of clear evaluation guidelines.

\subsubsection{Chain-of-Thought}
\label{subsubsec:meta_eval_cot}
We additionally investigate the impact of CoT on our evaluation by adding an explicit instruction (\eg, ``explain your reasoning before answering the question'') to encourage GPT to articulate its reasoning prior to delivering the final score. The complete prompt for evaluation is provided in \cref{subsec:supp_prompt_eval}. However, as shown in Table~\ref{tab:scoring_method}, our results indicate that CoT fails to enhance metric correlation consistently. In particular, CoT negatively affects the correlation for Perceptual Quality $PQ$ when using numeric scoring. 
Consequently, we utilize numeric scoring with detailed rubrics and CoT only for Instruction Following $IF$ and Identity Preservation $IP$, while CoT is disabled during Perceptual Quality $PQ$ evaluation.

\subsubsection{Instruction Input for Perceptual Quality}
\label{subsubsec:meta_eval_inst_input}
One might initially hypothesize that including the editing instructions alongside the output image could help the VLM discount aesthetically inferior modifications that are explicitly required by the instructions during Perceptual Quality $PQ$ evaluation. However, our experiments reveal that providing the editing instructions in this context dramatically reduces the correlation with human evaluations---from 0.234 to 0.046. Consequently, we exclude the editing instructions when evaluating Perceptual Quality $PQ$, ensuring that the VLM's assessments of image quality better align with human perception.

\subsubsection{Different Averaging Formula for Overall Score}
Previous studies~\cite{ku2024viescore,wei2025omniedit} advocate using the geometric mean to combine individual metrics to penalize low scores. In our case, switching from the geometric mean to the arithmetic mean resulted in a modest correlation increase from 0.334 to 0.386. Therefore, we compute the overall score as the arithmetic mean of the individual metrics.

\subsubsection{Comparison with CLIP Scores}
\label{subsubsec:meta_eval_clip}
We further compare the correlation of Directional CLIP Score $\text{CLIP}_\text{dir}$ and Image-wise CLIP Score $\text{CLIP}_\text{img}$ with human evaluation, as these scores similarly assess aspects of Instruction Following $IF$ and Identity
Preservation $IP$. As reported in \cref{tab:scoring_method}, these CLIP scores exhibit lower correlations with human evaluations compared to our metrics.

\begin{figure*}[t!]
    \centering
    \resizebox{\linewidth}{!}{
        \includegraphics[width=\textwidth]{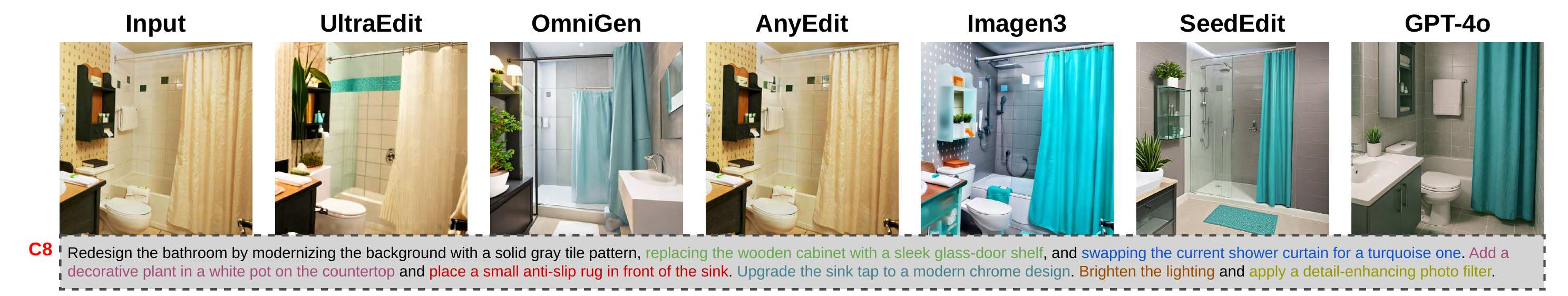}
    }
    \vspace{-2em}
    \caption{A real image edited with a $C_8$ instruction. Outputs from certain models tend to lose the realistic style completely.}
    \label{fig:vis_curse}
    \vspace{-1em}
\end{figure*}

\subsection{Per-Sample Variance}
\label{subsec:meta_eval_variance}

As discussed in \cref{subsec:metric_variance}, one way to reduce per-sample variance is to set the probability mass for sampling to an extremely small value, \eg, 1e-7, thereby approximating the determinism of proprietary VLMs. However, as shown in Table~\ref{tab:deterministic}, this approach slightly diminishes the correlation with human evaluations. Consequently, we disable determinstic sampling and instead perform multiple evaluations per sample, using the average score as the final metric.

To determine the required number of measurements per sample, we study the relationship between the number of measurements per sample and the variation in final scores across independent runs. Specifically, given $K$ measurements for one sample, we compute average scores four times, each averaged from $K$ distinct measurements, denoted as $S_1$, $S_2$, $S_3$, $S_4$. The variation for this sample is defined as $\frac{\max\{S_1,S_2,S_3,S_4\} - \min\{S_1,S_2,S_3,S_4\}}{2}$.

Figure~\ref{fig:variance} shows that the variation amplitudes for all three metrics converge when $K=20$. Therefore, we adopt 20 measurements per sample in all subsequent experiments.

\section{Experiments}
\label{sec:exp}

\subsection{Experiment Setup}
\paragraph{Dataset.} 
Our \texttt{Complex-Edit} dataset comprises both realistic and synthetic images. Specifically, we select 531 deduplicated images from the EMU-Edit test set~\cite{sheynin2024emu} to serve as our realistic image collection. In parallel, we use FLUX.1 \cite{flux2024} to generate an equivalent set of 531 synthetic images based on the captions associated with these real-world images, also sourced from EMU-Edit. Following \cref{sec:data}, we create editing instructions with the complexity ranging from $C_1$ to $C_8$ for these images.

\paragraph{Models.} Our experiments involve five advanced instruction-based image editing models: three open-source models (UltraEdit~\cite{zhao2024ultraedit}, OmniGen~\cite{xiao2024omnigen}, AnyEdit~\cite{yu2024anyedit}) and two proprietary models (Imagen3~\cite{baldridge2024imagen}, SeedEdit~\cite{shi2024seededit}). We evaluate each model on both realistic and synthetic images using editing instructions spanning complexity levels from $C_1$ to $C_8$. Note that, owing to usage policies for the proprietary models, results are reported on approximately 60\% of the images for Imagen3 and 95\% for SeedEdit.

Additionally, we perform a preliminary evaluation of the newly launched GPT-4o, which now features image generation capabilities. Without API access, we obtain output images via the GPT-4o web interface and center-crop them to match the input aspect ratio. Our evaluation of GPT-4o is limited to roughly 30\% of realistic images at the hardest instruction complexity level $C_8$.

\begin{table}[t!]
    \centering
    \resizebox{\linewidth}{!}{
        \begin{tabular}{@{}lcccccc@{}}
            \toprule
            Model & Prop. of Real Images & Realistic Style & IF & IP & PQ & O\\
            \midrule
            UltraEdit & \checkmark & \faStar & 6.56 & 5.93 & 7.29 & 6.59 \\
            OmniGen & \checkmark \checkmark \checkmark & \faStar \faStar \faStar \faStar \faStar & 6.25 & 6.42 & 7.54 & 6.74 \\
            AnyEdit & \checkmark \checkmark \checkmark & \faStar \faStar \faStar \faStar & 1.60 & \textbf{8.15} & 7.25 & 5.67 \\
            SeedEdit & \checkmark & \faStar \faStar & 8.49 & 6.91 & 8.74 & 8.04 \\
            Imagen3 & ? & \faStar & 7.56 & 6.55 & 7.67 & 7.26 \\
            GPT4o & ? & \faStar \faStar & \textbf{9.29} & 7.51 & \textbf{9.47} & \textbf{8.76} \\
            \bottomrule
        \end{tabular}
    }
    \vspace{-.8em}
    \caption{Evaluation results with direct editing on real images with the instruction complexity at $C_8$.}
    \vspace{-1em}
    \label{tab:curse_syn}
\end{table}

\subsection{Effect of Increasing Complexity}
\label{subsec:exp_main}

\begin{figure*}[t!]
    \centering
    \resizebox{\linewidth}{!}{
        \includegraphics[width=\textwidth]{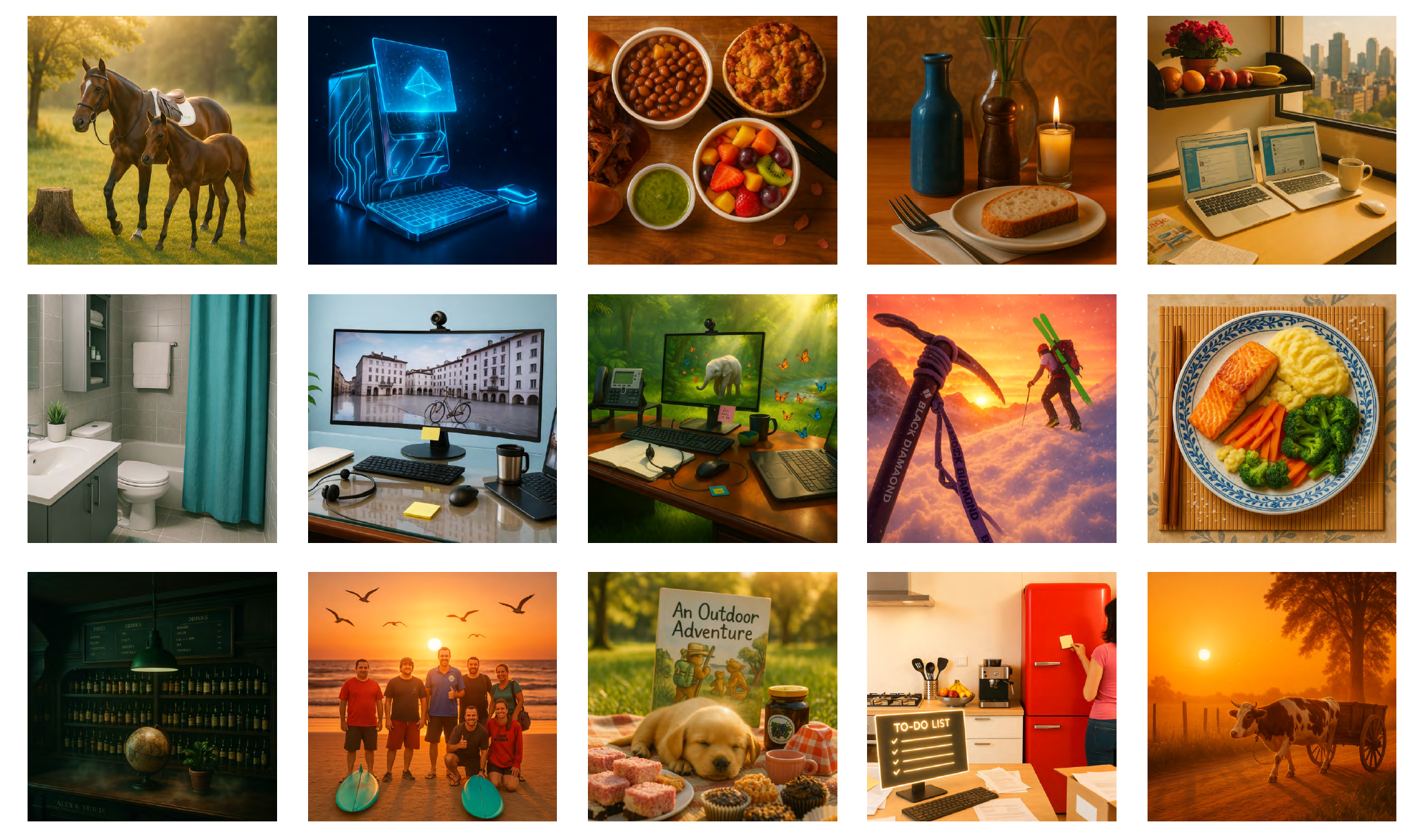}
    }
    \vspace{-2.6em}
    \caption{A real image edited with a $C_8$ instruction by GPT-4o. Outputs from GPT-4o severely lose the realistic style. See additional results in \cref{fig:vis_gpt}.}
    \label{fig:vis_main_gpt}
    \vspace{-1em}
\end{figure*}

\begin{table*}
    \centering
    \resizebox{.75\linewidth}{!}{
        \begin{tabular}{@{}lcccccccccccc@{}}
            \toprule
            \multirow{2}{*}{Model} & \multicolumn{4}{c}{$C_1$} & \multicolumn{4}{c}{$C_8$} & \multicolumn{4}{c}{$C_8 - C_1$} \\
            \cmidrule(lr){2-5} \cmidrule(lr){6-9} \cmidrule(lr){10-13}
            & $IF$ & $IP$ & $PQ$ & $O$ & $IF$ & $IP$ & $PQ$ & $O$ & $\Delta_{IF}$ & $\Delta_{IP}$ & $\Delta_{PQ}$ & $\Delta_{O}$\\
            \midrule
            UltraEdit & 7.13 & 7.76 & 7.45 & 7.45 & 6.56 & 5.93 & 7.29 & 6.59 & \perfdown{-0.57} & \perfdown{-1.82} & \perfdown{-0.16} & \perfdown{-0.85} \\
            OmniGen & 6.76 & 8.69 & 7.99 & 7.82 & 6.25 & 6.42 & 7.55 & 6.74 & \perfdown{-0.50} & \perfdown{-2.27} & \perfdown{-0.45} & \perfdown{-1.07} \\
            AnyEdit & 5.94 & 8.50 & 6.78 & 7.07 & 1.61 & \textbf{8.15} & 7.25 & 5.67 & \perfdown{-4.33} & \perfdown{-0.34} & \perfup{+0.47} & \perfdown{-1.40} \\
            Imagen3 & 7.67 & 8.93 & 7.90 & 8.17 & 7.56 & 6.55 & 7.68 & 7.27 & \perfdown{-0.11} & \perfdown{-2.38} & \perfdown{-0.22} & \perfdown{-0.90} \\
                SeedEdit & \textbf{9.31} & \textbf{9.01} & \textbf{8.71} & \textbf{9.01} & \textbf{8.50} & 6.91 & \textbf{8.74} & \textbf{8.05} & \perfdown{-0.81} & \perfdown{-2.10} & \perfup{+0.02} & \perfdown{-0.96} \\
            \bottomrule
        \end{tabular}
    }
    \vspace{-.8em}
    \caption{Performance comparison between $C_1$ and $C_8$ on real images. All models almost consistently underperform on all metrics when the complexity increases from $C_1$ to $C_8$.}
    \vspace{-.5em}
    \label{tab:main_result_real}
\end{table*}

\begin{table*}
    \centering
    \resizebox{.75\linewidth}{!}{
        \begin{tabular}{@{}lcccccccccccc@{}}
            \toprule
            \multirow{2}{*}{Model} & \multicolumn{4}{c}{$C_1$} & \multicolumn{4}{c}{$C_8$} & \multicolumn{4}{c}{$C_8 - C_1$} \\
            \cmidrule(lr){2-5} \cmidrule(lr){6-9} \cmidrule(lr){10-13}
            & $IF$ & $IP$ & $PQ$ & $O$ & $IF$ & $IP$ & $PQ$ & $O$ & $\Delta_{IF}$ & $\Delta_{IP}$ & $\Delta_{PQ}$ & $\Delta_{O}$\\
            \midrule
            UltraEdit & 7.82 & 8.61 & 9.02 & 8.48 & 6.51 & 6.83 & 8.72 & 7.35 & \perfdown{-1.30} & \perfdown{-1.78} & \perfdown{-0.30} & \perfdown{-1.13} \\
            OmniGen & 7.80 & 9.34 & 9.05 & 8.73 & 6.49 & 7.47 & 8.74 & 7.57 & \perfdown{-1.31} & \perfdown{-1.87} & \perfdown{-0.31} & \perfdown{-1.16} \\
            AnyEdit & 5.79 & 9.17 & 8.61 & 7.86 & 1.47 & \textbf{8.73} & 8.72 & 6.31 & \perfdown{-4.32} & \perfdown{-0.44} & \perfup{+0.11} & \perfdown{-1.55} \\
            Imagen3 & 8.27 & 9.25 & 8.86 & 8.79 & 7.97 & 7.54 & 8.88 & 8.13 & \perfdown{-0.30} & \perfdown{-1.72} & \perfup{+0.03} & \perfdown{-0.66} \\
            SeedEdit & \textbf{9.33} & \textbf{9.60} & \textbf{9.29} & \textbf{9.41} & \textbf{8.25} & 7.94 & \textbf{9.20} & \textbf{8.46} & \perfdown{-1.08} & \perfdown{-1.66} & \perfdown{-0.10} & \perfdown{-0.95} \\
            \bottomrule
        \end{tabular}
    }
    \vspace{-.8em}
    \caption{Performance comparison between $C_1$ and $C_8$ on synthetic images. All models almost consistently underperform on all metrics when the complexity increases from $C_1$ to $C_8$.}
    \vspace{-1.3em}
    \label{tab:main_result_syn}
\end{table*}

\begin{figure*}[h!]
    \centering
    \begin{subfigure}{\linewidth}
        \centering
        \includegraphics[width=.99\linewidth]{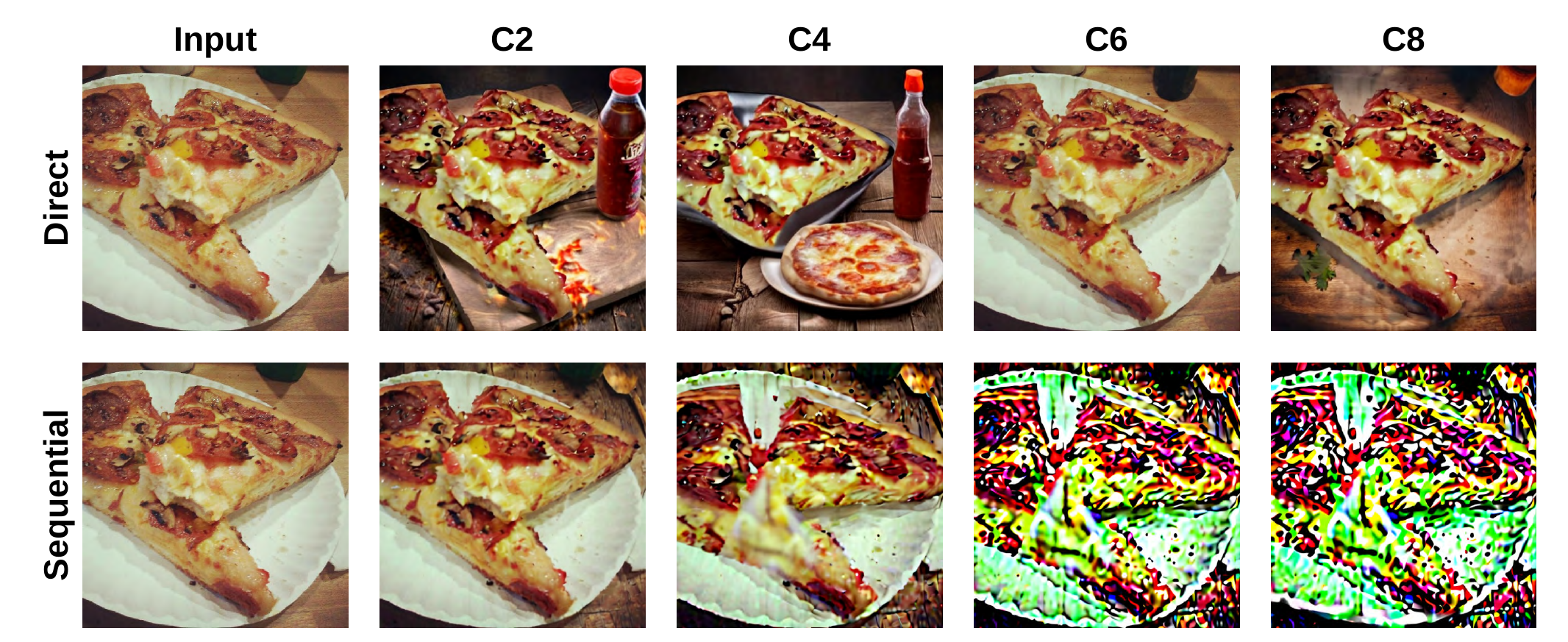}
        \vspace{-.5em}
        \caption{real input images.}
    \label{fig:vis_direct_sequence_real}
    \end{subfigure}
    \hfill
    \vspace{-2em}
    \begin{subfigure}{\linewidth}
        \centering
        \includegraphics[width=.99\linewidth]{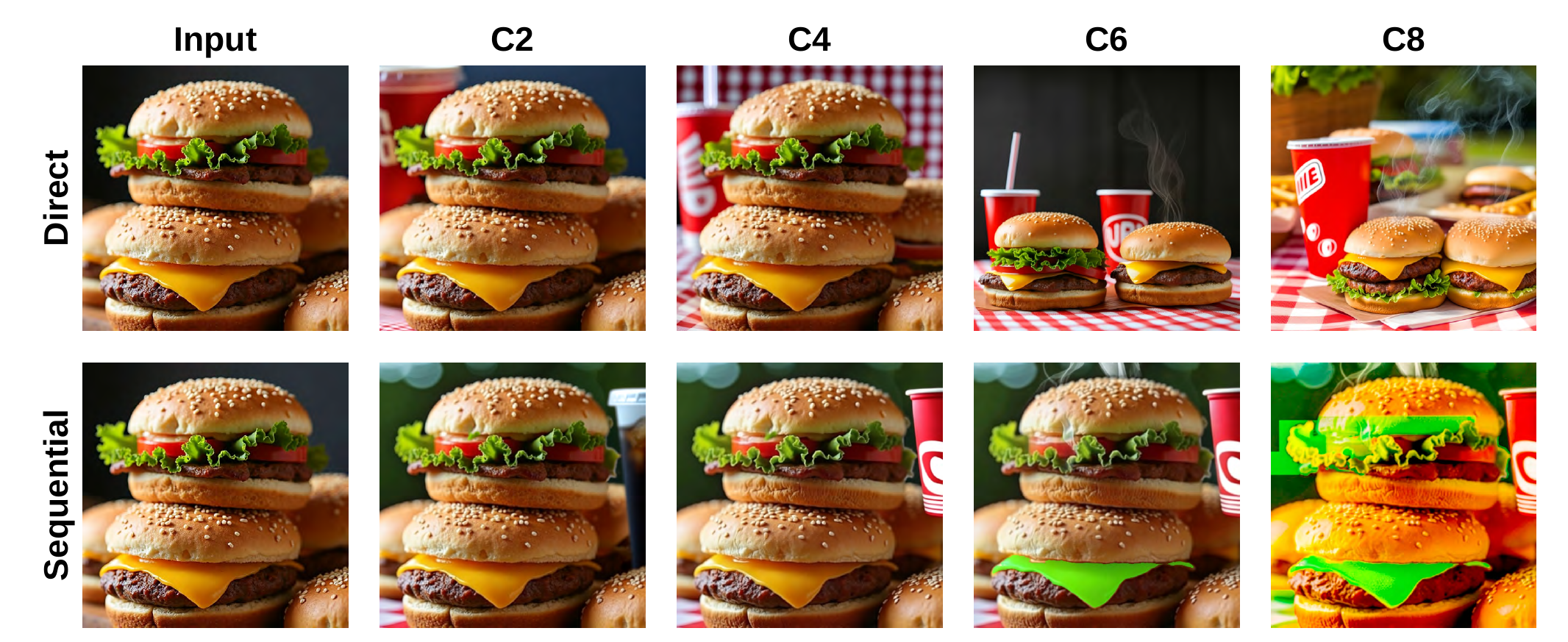}
        \vspace{-.5em}
        \caption{Synthetic input images.}
        \label{fig:vis_direct_sequence_syn}
    \end{subfigure}
    \vspace{-2em}
    \caption{Qualitative results of direct and sequential editing. The complexity level ranges from $C_1$ to $C_8$. It can be seen that $IF$ and $PQ$ are severely hurt by the growing instruction complexity. The instructions are the same ones in \cref{fig:vis_main}. Qualitative results with more models are shown in the supplementary~\cref{fig:vis_direct_real,fig:vis_direct_syn,fig:vis_sequence_real,fig:vis_sequence_syn}.}
    \label{fig:vis_direct_sequence}
    \vspace{-1em}
\end{figure*}

\begin{figure*}[ht]
    \centering
    \resizebox{\linewidth}{!}{
        \includegraphics[width=\linewidth]{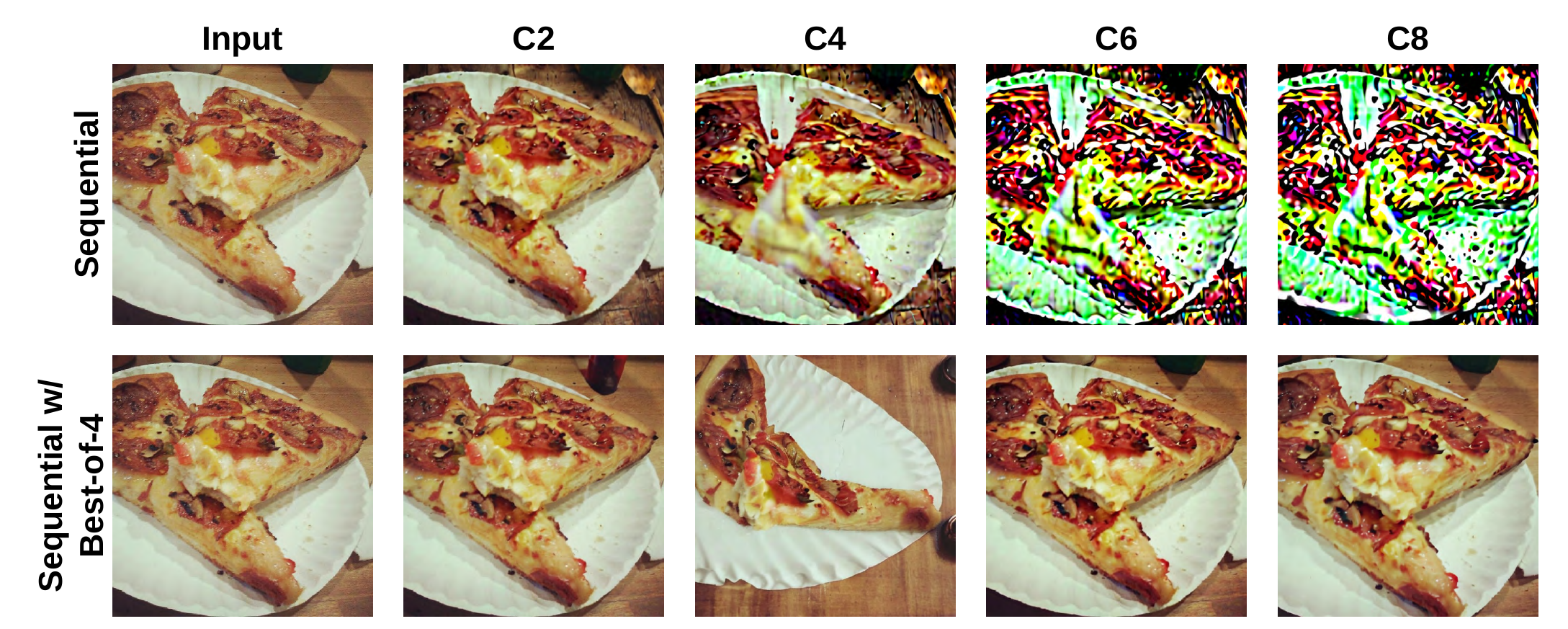}
    }
    \vspace{-2em}
    \caption{Qualitative results of sequential editing with and without Best-of-$4$ with OmniGen on real input images. The complexity level is at $C_8$. This shows that sequential editing can benefit a lot from Best-of-$N$ in terms of $IF$ and $PQ$. The instructions are the same ones in \cref{fig:vis_main_real}. Qualitative results with more models are shown in the supplementary~\cref{fig:vis_sequence_best4_real}.}
    \label{fig:vis_sequence_best1_best4_real}
\end{figure*}

\begin{figure*}[ht]
    \centering
    \resizebox{\linewidth}{!}{
        \includegraphics[width=\linewidth]{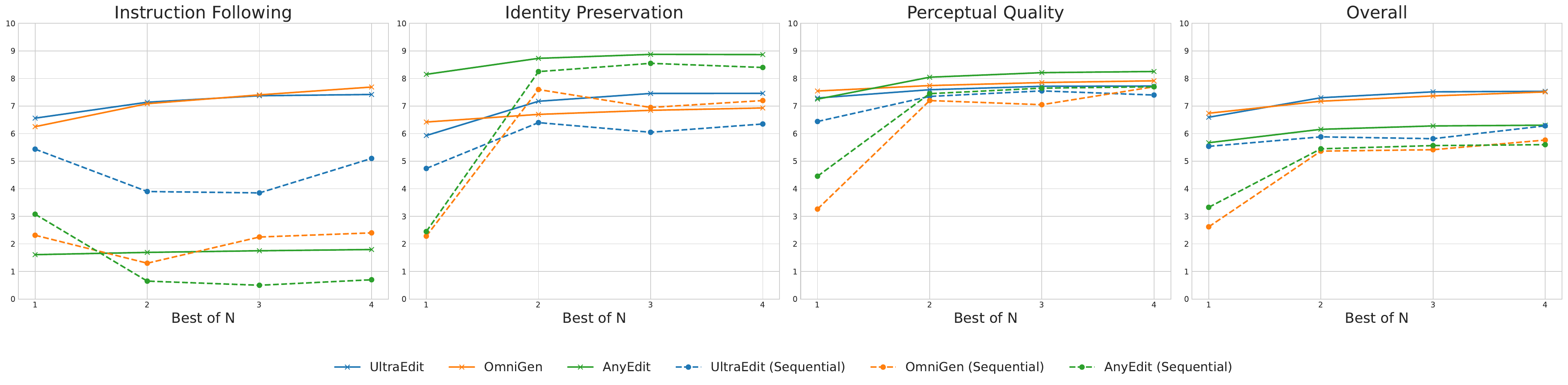}
    }
    \vspace{-2em}
    \caption{Direct and Sequential Editing results w/ Best-of-$N$ of open-source models on real input images. The complexity level is at $C_8$. This shows that Best-of-$N$ can improve $IF$ and $PQ$ especially for sequential editing. However, sequential editing with Best-of-$N$ can still barely surpass direct editing without Best-of-$N$.}
    \label{fig:bestn_eval}
\end{figure*}

Figure~\ref{fig:main_eval} illustrates the overall performance trends as instruction complexity increases. In general, higher complexity leads to a consistent decline in both Identity Preservation and Perceptual Quality across models, while Instruction Following tends to fluctuate depending on the specific model. For instance, although AnyEdit exhibits a substantial drop in Instruction Following as complexity increases, its Perceptual Quality improves moderately. Additionally, proprietary models outperform open-source alternatives on both realistic and synthetic images, and the performance gap becomes more pronounced at higher complexity levels. Furthermore, as shown in \cref{tab:main_result_real,tab:main_result_syn}, open-source models' performance drops are notably more severe than proprietary models. For example, when editing synthetic images, all open-source models have drops larger than 1 in Overall score while the Overall score's drop of Imagen3 and SeedEdit are $0.66$ and $0.95$ respectively, implying that stronger models are not more resilient towards the negative impact of increased instruction complexity. Sample qualitative results are shown in Figure~\ref{fig:vis_main}.

Our preliminary evaluation of GPT-4o, as detailed in Table~\ref{tab:curse_syn}, demonstrates that it significantly outperforms all other models at the hardest instruction complexity level $C_8$, particularly in terms of Instruction Following and Perceptual Quality.

\subsection{Curse of Synthetic data}
\label{subsec:exp_curse}
Our qualitative analysis reveals an intriguing phenomenon: when applying extremely complex editing instructions ($C_8$) to real input images, the resulting outputs frequently lose their realistic appearance and adopt a distinctly synthetic aesthetic. Notably, UltraEdit is particularly susceptible to this effect compared to OmniGen and AnyEdit, as illustrated in Figure~\ref{fig:vis_curse}. We attribute this trend to the composition of UltraEdit’s training data, which contains a significantly higher proportion of synthetic images than the datasets used by OmniGen and AnyEdit (see Table~\ref{tab:curse_syn}). This observation aligns with the ``curse of synthetic data'' phenomenon discussed in~\citep{gerstgrasser2024model}.

This phenomenon also extends to highly capable proprietary models, including SeedEdit, Imagen3, and even GPT-4o. Specifically, as illustrated in Figure~\ref{fig:vis_main_gpt}, the edited images produced by GPT-4o completely lose their realistic style, even though its overall performance surpasses that of all other models evaluated. This observation may indicate that, although the training sources for Imagen3 and GPT-4o are undisclosed, these models likely incorporate synthetic data into their training sets, which in turn contributes to their impressive performance as well as the emergence of a distinctly synthetic aesthetic.

\subsection{Test-Time Scaling Approaches}
\label{subsec:exp_tts}

\subsubsection{CoT-like Sequential Editing}
\label{subsubsec:exp_sequential}
Our data pipeline composes complex instructions by combining sequences of atomic editing operations, suggesting that sequential application of these atomic steps should be equivalent to executing the complex instruction directly. Moreover, this way of decomposing a complex task and then sequentially executing it has been proven effective in both language generation~\cite{wei2022chain} and text-to-image generation~\cite{guo2025can}.

Formally, given an input image $x$, a complex instruction $T$ at complexity level $C_i$, and a corresponding sequence of atomic instructions ${t_1, \dots, t_i}$, we define the outputs for direct and sequential editing as follows:
\begin{align}
    y_\text{direct} &= f(x, T) \\
    y_{\text{sequence}} &= f^i(x, \{t_1, \dots, t_i\}) \\
    &= f(f(f(\dots f(x, t_1) \dots, t_{i-1}), t_i))
\end{align}

We evaluate sequential editing on both real and synthetic images. Our results (summarized in Figure~\ref{fig:main_eval} and illustrated qualitatively in Figure~\ref{fig:vis_direct_sequence}) reveal that sequential editing yields a steady decline in performance across all three metrics, with accumulating visual artifacts and distortions---even for strong proprietary models such as Imagen3 and SeedEdit. Notably, AnyEdit demonstrates improved Instruction Following with sequential editing compared to direct editing, although this improvement is offset by a reduction in Identity Preservation.

\subsubsection{Best-of-$N$}
\label{subsubsec:exp_bestn}
We also experiment with Best-of-N, which is another simple test-time-scaling method commonly used for text-to-image generation~\cite{guo2025can,ma2025inference}.
For direct editing, we generate $N$ candidate outputs for each input image and select the one with the highest overall score $O$. For sequential editing, we generate $N$ candidates at each step and choose the candidate with the highest overall score $O$ to proceed to the next editing operation. When evaluating an intermediate candidate at the $i$-th step during sequential editing, instead of considering it as the output of a single-step edit produced with the predecessor output and an atomic instruction, \eg $f(y_{i-1}, t_i)$ with $y_{i-1} = f^{i-1}(x, \{t_1, \dots, t_{i-1}\})$, we evaluate it as the output from the original input image with a compound instruction $T_i$. This compound instruction $T_i$ has $C_i$ complexity and encompasses all the atomic instructions up to the $i$-th step: $\{t_1, \dots, t_i\}$. The rationale here is that the intermediate result should best represent the cumulative effects of all preceding atomic steps on the original image, ensuring the final output best reflects all atomic steps and consequently the final compound instruction.

Our evaluations with open-source models on real-life images (see Figure~\ref{fig:bestn_eval}) indicate that increasing $N$ gradually improves direct editing performance across all metrics. For sequential editing, even a Best-of-$N$ strategy with $N=2$ produces significant gains in Identity Preservation and Perceptual Quality; however, the improvement in Instruction Following is less consistent. Qualitative comparisons are provided in Figure~\ref{fig:vis_sequence_best1_best4_real}.

\section{Related Works}
\label{sec:related_works}

\paragraph{Instruction-Based Image Editing}
InstructPix2Pix~\cite{brooks2023instructpix2pix} established the paradigm of instruction-based image editing, whereby images are directly modified through textual instructions, thus enhancing user interaction efficiency. This approach necessitated the creation of a substantial dataset for instruction-based image editing, constructed through the utilization of fine-tuned GPT-3~\cite{achiam2023gpt} and the Prompt-to-Prompt~\cite{hertz2022prompt} Diffusion Model. Nevertheless, the editing capabilities were constrained by the limited generation quality of these underlying models. Subsequent research endeavors have sought to address these limitations through various methodological innovations. MagicBrush~\cite{kawar2023imagic} proposed a manually curated annotation methodology, while HQ-Edit\cite{hui2025hqedit} employed sophisticated black-box large language models in conjunction with text-to-image generation systems to produce datasets of superior quality. To enhance instruction granularity, MGIE~\cite{fu2023guiding} and SmartEdit~\cite{huang2024smartedit} implemented multilevel language learning models (MLLLM), facilitating more precise editing operations. Furthermore, UltraEdit~\cite{zhao2024ultraedit} introduced advanced masking techniques to enable fine-grained image manipulation. OmniGen~\cite{xiao2024omnigen} expanded both the model architecture and data corpus, implementing a unified Transformer framework capable of concurrent processing and comprehension of textual and visual inputs while inferring editing targets from linguistic descriptions. UIP2P~\cite{simsar2024uip2p} introduced an innovative cyclic editing methodology, enabling unsupervised instruction-driven image editing without the requirement for paired edited images during the training phase. The domain of instruction-based image editing has witnessed substantial advancements through these diverse methodological contributions.

\paragraph{Test-Time Scaling for Image Generation}
Recently, inspired by Large Reasoning models such as OpenAI's GPT4-o and DeepSeek-R1, substantial efforts have been dedicated to enhancing image generation results through Test-Time Scaling methodologies.  \citet{ma2025inference} employed verifiers to provide critical feedback while utilizing various search algorithms to identify optimal noise parameters, thereby enhancing the quality of generated samples during the inference stage. \citet{xie2025sana} implemented Repeated Sampling techniques in conjunction with VLM-based Evaluation mechanisms to improve both the overall aesthetic quality of generated images and their semantic alignment with textual prompts. \citet{guo2025can} introduced test-time verification and preference alignment within the autoregressive image generation process, significantly improving the quality of text-to-image generation. Although \citet{guo2025can} have successfully implemented CoT-like sequential generation in text-to-image generation tasks, maintaining image identity remains a critical challenge in image editing contexts, imposing substantially higher constraints and requirements on editing models. The preservation of original image elements while executing targeted modifications necessitates more sophisticated approaches than those employed in other less-constrained image generation tasks, thus presenting a distinctive set of technical challenges for instruction-based editing systems.

In this paper, rather than proposing more powerful instruction-based image editing models, we introduce a challenging instruction editing dataset designed to systematically assess how existing models perform under complex instructions.

\section{Conclusion}
In this work, we introduced \texttt{Complex-Edit}, a comprehensive benchmark designed to systematically evaluate instruction-based image editing models across varying levels of instruction complexity. Through extensive experiments, \texttt{Complex-Edit} enable us to uncover multiple key insights regarding the limitations and capabilities of current models.
We hope this benchmark will drive future advancements in developing more powerful instruction-based image editing models, especially those with test-time scaling ability.

\section*{Acknowledgement}
We would like to thank Google Cloud Research Credits Program, and the Microsoft Accelerate
Foundation Models Research Program for supporting our computing needs.

{
    \small
    \bibliographystyle{ieeenat_fullname}
    \bibliography{main}
}

\clearpage
\setcounter{page}{1}
\maketitlesupplementary

\section{Atomic Editing Operations}
\label{sec:supp_atomic}
Here we present the descriptions for all the 24 atomic operation types. These operation types are also fed to GPT-4o during Stage \#1 (Sequence Generation) of our data pipeline, as discussed in \cref{subsec:data_sequence}.

\begin{itemize}
    \item \textbf{Object Manipulation and Transformation}
    \begin{itemize}
        \item \textit{Add an Object}: Insert a new element into the image.
        \item \textit{Remove an Object}: Eliminate an existing element from the image.
        \item \textit{Replace an Object}: Swap one element with another.
        \item \textit{Move an Object}: Change the position of an existing element within the image.
        \item \textit{Resize an Object}: Adjust the size of an existing element.
        \item \textit{Rotate an Object}: Rotate an element to a specified angle.
        \item \textit{Duplicate an Object}: Create a copy of an existing element.
    \end{itemize}
    
    \item \textbf{Color and Tone Adjustments}
    \begin{itemize}
        \item \textit{Change Color}: Replace the color of an element with a specified color.
        \item \textit{Apply Filter/Weather}: Add a color filter or weather effect to the entire image or specific parts.
    \end{itemize}
    \item \textbf{Texture and Material Adjustments}
    \begin{itemize}
        \item \textit{Change Texture}: Apply a texture to an element (e.g., change from metal to wood).
    \end{itemize}
    
    \item \textbf{Background and Environment}
    \begin{itemize}
        \item \textit{Change Background}: Replace the background with a different scene or color.
    \end{itemize}
    \item \textbf{Lighting and Shadows}
    \begin{itemize}
        \item \textit{Adjust Lighting}: Change the overall lighting or lighting of specific elements.
    \end{itemize}
    \item \textbf{Text and Symbols}
    \begin{itemize}
        \item \textit{Add Text}: Insert text into the image.
        \item \textit{Remove Text}: Eliminate existing text from the image.
        \item \textit{Change Text Properties}: Modify font, color, size, or position of existing text.
    \end{itemize}
    \item \textbf{Pose and Expression}
    \begin{itemize}
        \item \textit{Change Pose}: Modify the stance or posture of a person or object.
        \item \textit{Change Facial Expression}: Alter the facial expression of a character.
    \end{itemize}
    \item \textbf{Composition and Cropping}
    \begin{itemize}
        \item \textit{Crop Image}: Adjust the framing of the image by removing outer areas.
        \item \textit{Reframe Composition}: Change the focus or arrangement of elements within the image.
        \item \textit{Zoom In/Out}: Adjust the zoom level to focus on specific elements or show a broader view.
    \end{itemize}
    \item \textbf{Special Effects}
    \begin{itemize}
        \item \textit{Add Special Effects}: Introduce effects like glow, motion blur, or lens flare.
        \item \textit{Remove Special Effects}: Eliminate existing special effects from the image.
        \item \textit{Add Particles}: Insert particles like dust.
        \item \textit{Remove Particles}: Remove existing particles from the image.
    \end{itemize}
\end{itemize}

\section{Prompts for Evaluation}
\subsection{Rubrics}
\label{subsec:supp_prompt_rubrics}

In the rubric for each metric we propose, we specify the detailed textual description for every score from 0 to 10, and most scores have a textual example to reduce ambiguity. All the rubrics for evaluation in are listed in \cref{fig:rubric}.

\begin{figure*}[ht]
    \centering
    \begin{subfigure}{0.99\linewidth}
        \centering
        \input{tcolorbox/rubric/if}
        \caption{Instruction Following.}
        \label{fig:rubric_if}
    \end{subfigure}
    \hfill
    \begin{subfigure}{0.99\linewidth}
        \centering
        \input{tcolorbox/rubric/ip}
        \caption{Identity Preservation.}
        \label{fig:rubric_ip}
    \end{subfigure}
    \hfill
    \begin{subfigure}{0.99\linewidth}
        \centering
        \input{tcolorbox/rubric/pq}
        \caption{Perceptual Quality.}
        \label{fig:rubric_pq}
    \end{subfigure}
    \caption{Rubrics for evaluating Instruction Following, Identity Preservation and Perceptual Quality.}
    \label{fig:rubric}
\end{figure*}

\subsection{System Prompts}
\label{subsec:supp_prompt_eval}
The system prompts that we use for evaluation are provided in \cref{fig:system_prompt}.

\begin{figure*}[ht]
    \centering
    \begin{subfigure}{0.99\linewidth}
        \centering
        \input{tcolorbox/system/alignment}
        \caption{Instruction Following and Identity Preservation.}
        \label{fig:system_prompt_alignment}
    \end{subfigure}
    \hfill
    \begin{subfigure}{0.99\linewidth}
        \centering
        \input{tcolorbox/system/quality}
        \caption{Perceptual Quality.}
        \label{fig:system_prompt_quality}
    \end{subfigure}
    \caption{System prompt for evaluating Instruction Following, Identity Preservation and Perceptual Quality.}
    \label{fig:system_prompt}
\end{figure*}

\section{Additional Qualitative Results}
Here we present more comprehensive qualitative results.

\subsection{Direct Editing}
\begin{figure*}[ht]
    \centering
    \resizebox{\linewidth}{!}{
        \includegraphics[width=0.99\linewidth]{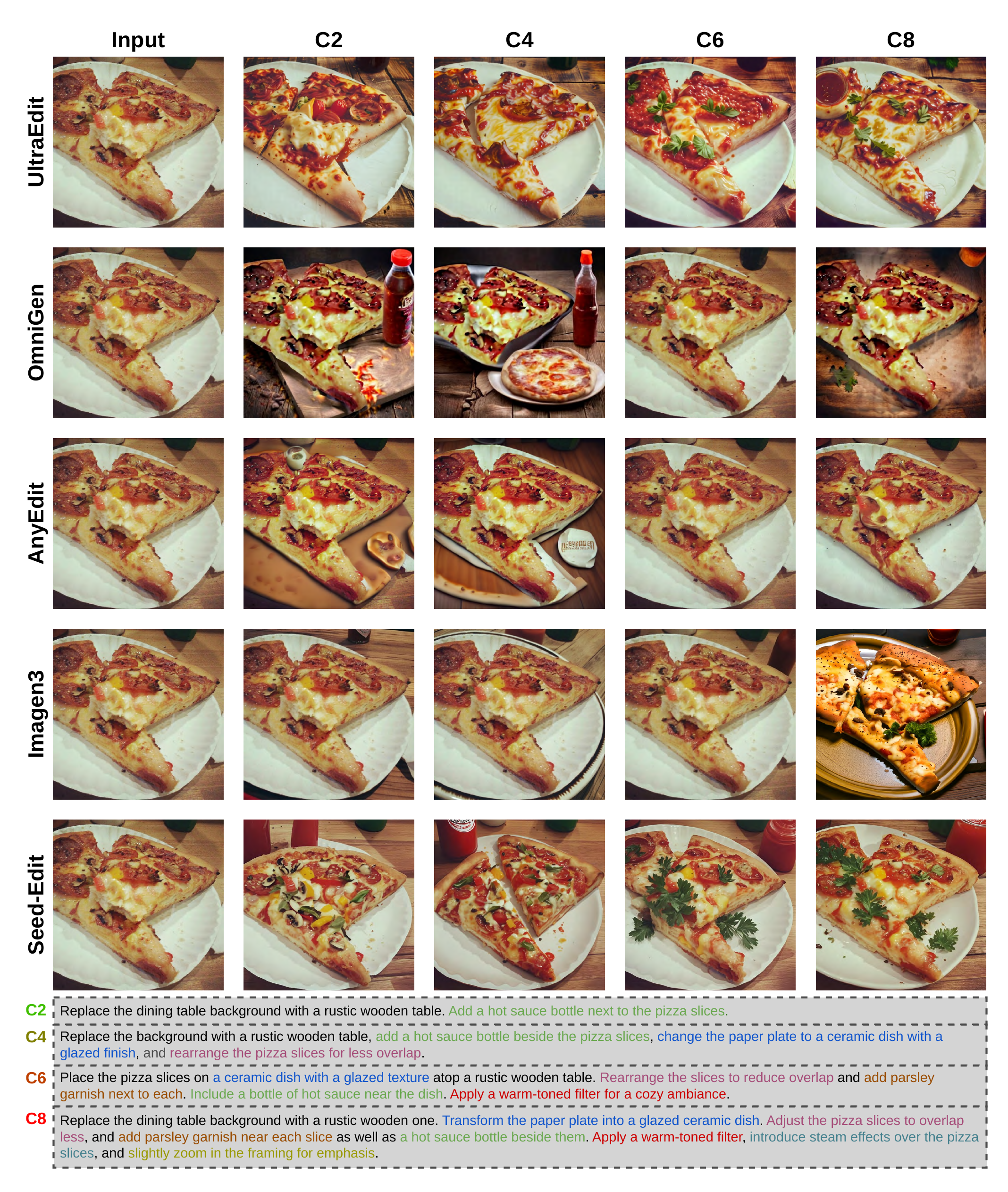}
    }
    \vspace{-2em}
    \caption{Additional qualitative results with direct editing on real-life images. As instruction complexity increases, the gap between open-source and proprietary models widens, and both Identity Preservation and Perceptual Quality decline, with variable effects on Instruction Following across models. See \cref{subsec:exp_main} for more details.}
    \label{fig:vis_direct_real}
\end{figure*}

\begin{figure*}[ht]
    \centering
    \resizebox{\linewidth}{!}{
        \includegraphics[width=0.99\linewidth]{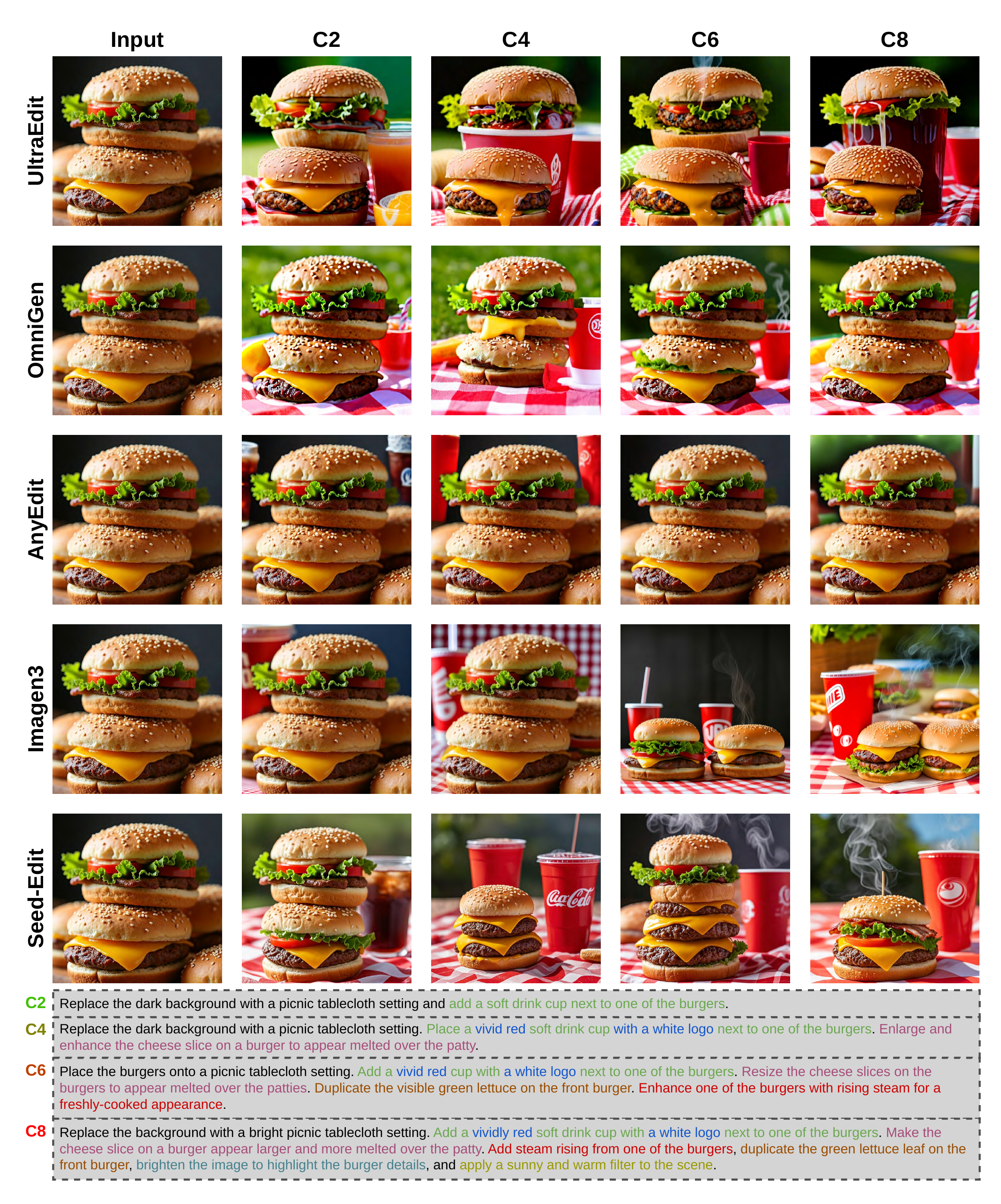}
    }
    \vspace{-2em}
    \caption{Additional qualitative results with direct editing on synthetic images. As instruction complexity increases, the gap between open-source and proprietary models widens, and both Identity Preservation and Perceptual Quality decline, with variable effects on Instruction Following across models. See \cref{subsec:exp_main} for more details.}
    \label{fig:vis_direct_syn}
\end{figure*}

Additional qualitative results of direct editing with more models are shown in \cref{fig:vis_direct_real,fig:vis_direct_syn}, indicating that the gap between open-source and proprietary models expands with increasing complexity of the instruction. Also, Identity Preservation and Perceptual Quality continue to drop as instruction complexity increases, while the effect on Instruction Following varies from model to model. One can refer to the discussion in \cref{subsec:exp_main} for more detailed insights.

\subsection{Sequential Editing}
\begin{figure*}[ht]
    \centering
    \resizebox{\linewidth}{!}{
        \includegraphics[width=0.99\linewidth]{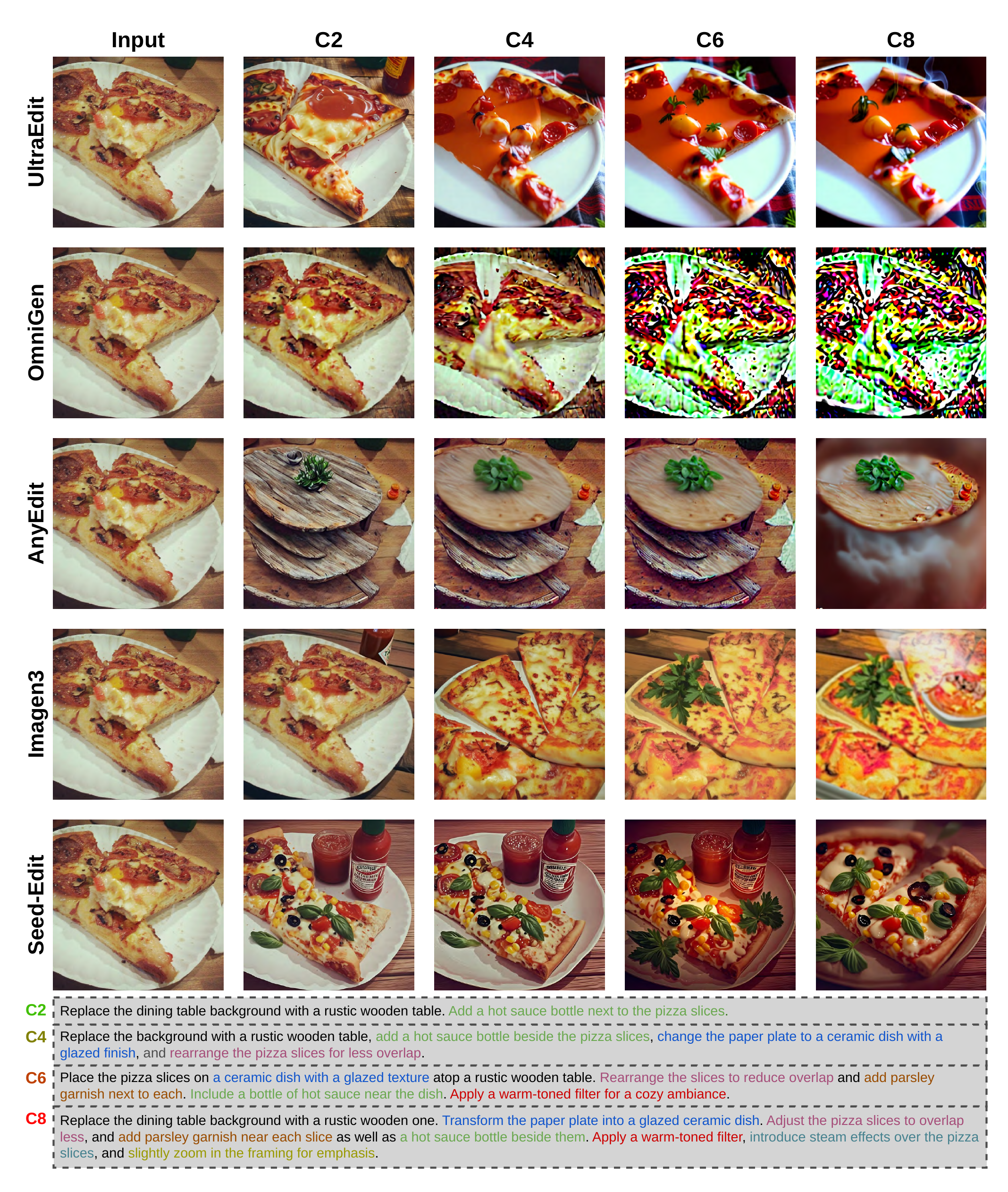}
    }
    \vspace{-2em}
    \caption{Additional qualitative results with sequential editing on real-life images. Identity Preservation and Perceptual Quality, degrade as visual artifacts and distortions increase with more intermediate steps. Even advanced proprietary models, \ie Imagen3 and SeedEdit, struggle to maintain element identity and quality in output images. More in-depth analysis is in \cref{subsubsec:exp_sequential}.}
    \label{fig:vis_sequence_real}
\end{figure*}

\begin{figure*}[ht]
    \centering
    \resizebox{\linewidth}{!}{
        \includegraphics[width=0.99\linewidth]{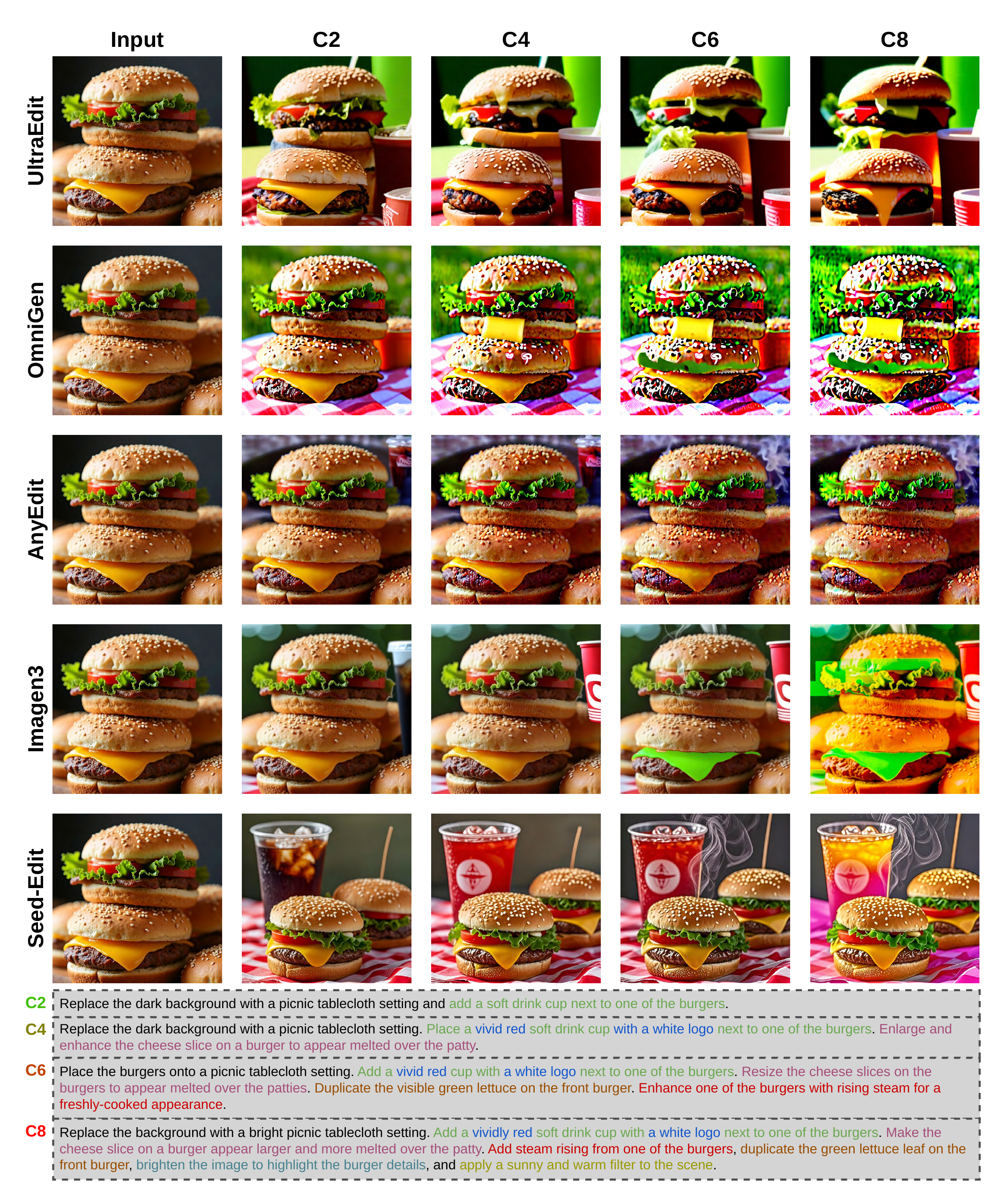}
    }
    \vspace{-2em}
    \caption{Additional qualitative results with sequential editing on synthetic images. Identity Preservation and Perceptual Quality, degrade as visual artifacts and distortions increase with more intermediate steps. Even advanced proprietary models, \ie Imagen3 and SeedEdit, struggle to maintain element identity and quality in output images. More in-depth analysis is in \cref{subsubsec:exp_sequential}.}
    \label{fig:vis_sequence_syn}
\end{figure*}

We illustrate additional qualitative results of sequential editing with more models in \cref{fig:vis_sequence_real,fig:vis_sequence_syn}, showing that all three aspects, particularly Identity Preservation and Perceptual Quality, consistently degrade, as visual artifacts and distortions accumulate with an increase in the number of intermediate steps. Even advanced proprietary models, \ie Imagen3 and SeedEdit, demonstrate limitations in maintaining the identity of key elements and the overall quality of output images. More in-depth discussions are given in \cref{subsubsec:exp_sequential}.

\subsection{Best-of-N}
\begin{figure*}[ht]
    \centering
    \resizebox{\linewidth}{!}{
        \includegraphics[width=0.99\linewidth]{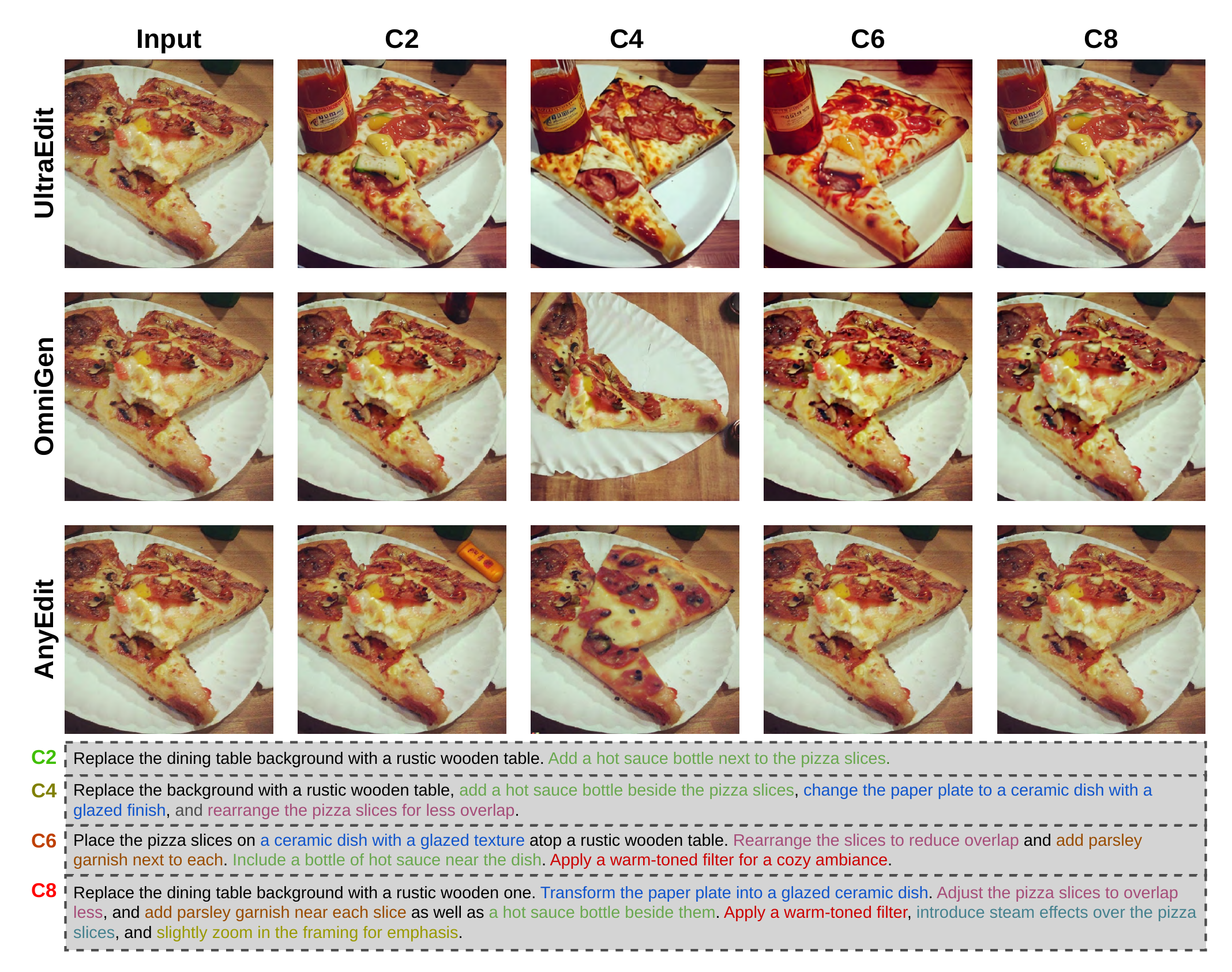}
    }
    \vspace{-2em}
    \caption{Qualitative results with sequential editing with Best-of-4 on real-life images. Identity Preservation and Perceptual Quality of sequential editing improve significantly with Best-of-4. Refer to \cref{subsubsec:exp_bestn} for more detailed discussions.}
    \label{fig:vis_sequence_best4_real}
\end{figure*}

Additional qualitative results of sequential editing combined with Best-of-4 with more open-source models are shown in \cref{fig:vis_sequence_best4_real}. It can been noticed that the Identity Preservation and Perceptual Quality of sequential editing's results can be substantially improved with Best-of-4. A more comprehensive analysis of this topic is provided in \cref{subsubsec:exp_bestn}. 

The Identity Preservation and Perceptual Quality of sequential editing improve significantly with Best-of-4, as analyzed in detail in \cref{subsubsec:exp_bestn}.

\subsection{Direct Editing with GPT-4o}
\begin{figure*}[t!]
    \centering
    \resizebox{\linewidth}{!}{
        \includegraphics[width=0.99\textwidth]{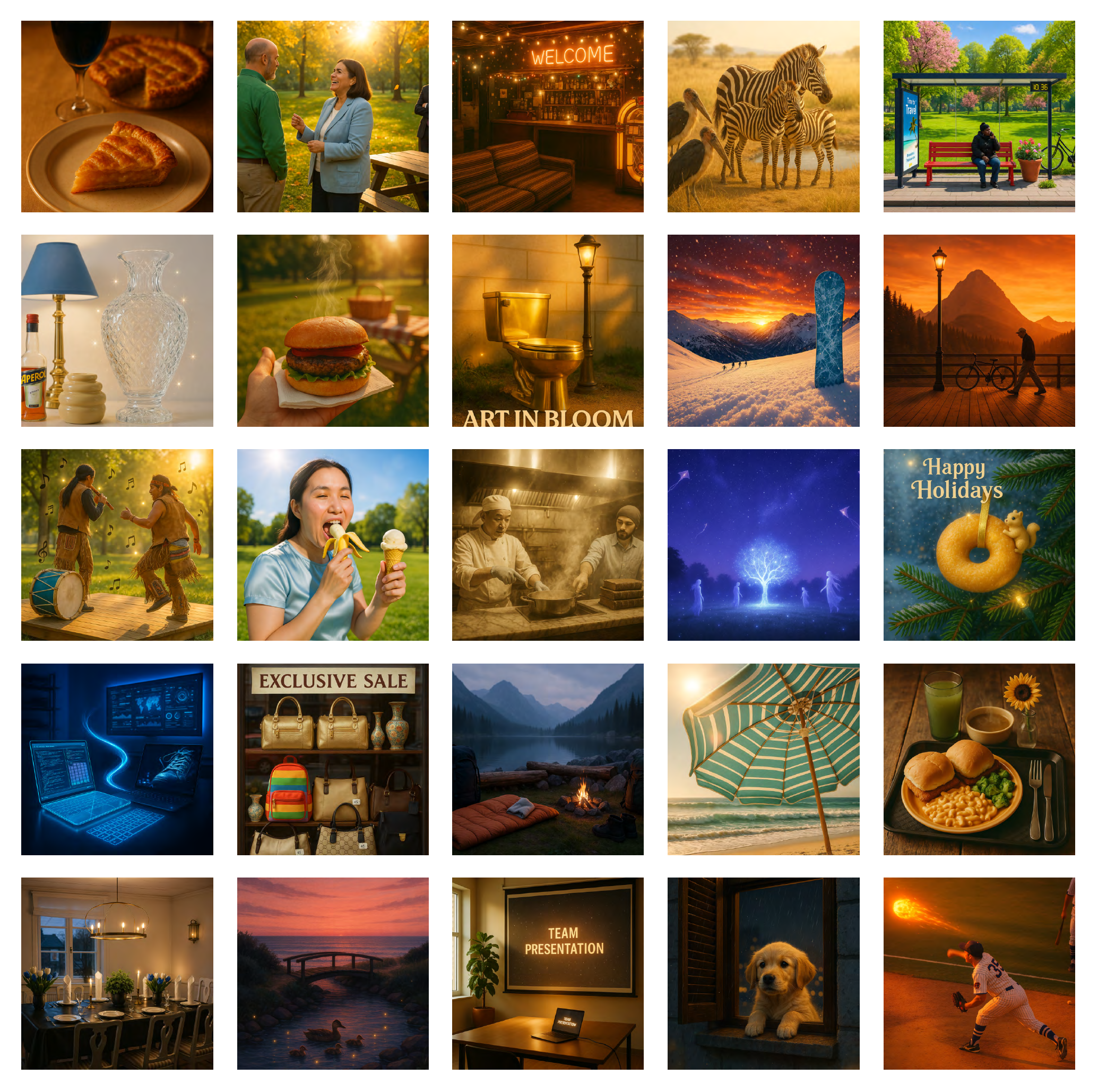}
    }
    \vspace{-2em}
    \caption{More real-life images edited with a $C_8$ instruction by GPT-4o. Outputs from GPT-4o severely lose the realistic style. Refer to \cref{subsec:exp_curse} for a detailed discussion.}
    \label{fig:vis_gpt}
\end{figure*}

We provide more qualitative results with GPT-4o via direct editing real-life images in \cref{fig:vis_gpt}. These results imply that while GPT-4o is capable of generating edited images of exceptional quality, output images with very complex instructions prune to lose the realistic feeling in the original images. This may imply a lack of real-life images in GPT-4o's training data. See \cref{subsec:exp_curse} for a more in-depth analysis.

\end{document}